\documentclass{article}
\usepackage[utf8]{inputenc}  % no template original é latin1, espero q nao de problema
\usepackage[T1]{fontenc}
\usepackage[brazil,english]{babel}   
\usepackage{graphicx,xurl,color}
\usepackage{hyperref}
\usepackage{breakcites}
\usepackage{todonotes}

\newcommand{\rew}{recompensa}
\newcommand{\rl}{AR} 
\newcommand{\ag}{agente}
\newcommand{\ql}{Q-learning}

\author{Ana L.C. Bazzan, Anderson R. Tavares, André G. Pereira,\\ Cláudio R. Jung, Jacob Scharcanski, Joel Luis Carbonera, Luís C. Lamb,\\ Mariana Recamonde-Mendoza, Thiago L.T. da Silveira, Viviane Moreira\footnote{Authors in first-name alphabetical order; equal contribution. All with the Institute of Informatics, UFRGS, Porto Alegre, Brazil. \emph{pt} Lista de autores em ordem alfabética. Contribuições iguais. Todos autores são afiliados ao Instituto de Informática, Universidade Federal do Rio Grande do Sul, Porto Alegre, Brasil.}}
\title{``A Nova Eletricidade'': \\ Aplicações, Riscos e Tendências da IA Moderna\footnote{``The New Electricity'': Applications, Risks, and Trends in Current AI}}

\begin{document} 

\maketitle

\begin{abstract}
The thought-provoking analogy between AI and electricity, made by computer scientist and entrepreneur Andrew Ng, summarizes the deep transformation that recent advances in Artificial Intelligence (AI) have triggered in the world. 
This chapter presents an overview of the ever-evolving landscape of AI, written in Portuguese. With no intent to exhaust the subject, we explore the AI applications that are redefining sectors of the economy, impacting society and humanity. We analyze the risks that may come along with rapid technological progress and future trends in AI, an area that is on the path to becoming a general-purpose technology, just like electricity, which revolutionized society in the 19th and 20th centuries.

\begin{center}{\textbf{Resumo}}\end{center}
A provocativa comparação entre IA e eletricidade, feita pelo cientista da computação e empreendedor Andrew Ng, resume a profunda transformação que os recentes avanços em Inteligência Artificial (IA) têm desencadeado no mundo. 
Este capítulo apresenta uma visão geral pela paisagem em constante evolução da IA. Sem pretensões de exaurir o assunto, exploramos as aplicações que estão redefinindo setores da economia, impactando a sociedade e a humanidade. Analisamos os riscos que acompanham o rápido progresso tecnológico e as tendências futuras da IA, área que trilha o caminho para se tornar uma tecnologia de propósito geral, assim como a eletricidade, que revolucionou a sociedade dos séculos XIX e XX. 

\end{abstract}

%\part*{Parte I: Introdução e Fundamentos de IA}
\part*{Parte I: Introdução e Fundamentos}

\section{Introdução}
\label{sec:intro}
Comparar a Inteligência Artificial (IA) e a eletricidade foi a maneira que o cientista da computação e empreendedor Andrew Ng usou para sintetizar o potencial transformador e também os perigos dessa tecnologia \cite{Lynch2017andrewng}.  Assim como a eletricidade moldou a história do século XIX, a IA está esculpindo o cenário do século XXI de maneiras que desafiam as fronteiras do conhecimento e da imaginação, com um farol de possibilidades intrigantes e, ao mesmo tempo, profundas preocupações sobre os rumos que a tecnologia pode tomar, mesmo quando usada para fins não-maliciosos.

Indo além da analogia IA-eletricidade de Andrew Ng, o status da IA como uma força transformadora foi reconhecido pela comunidade científica da Ciência da Computação. O prestigiado Prêmio Turing, também chamado de ``Nobel da Computação'', foi concedido em 2018 aos cientistas da computação Yoshua Bengio, Geoffrey Hinton e Yann LeCun por suas contribuições pioneiras para o desenvolvimento da aprendizagem profunda, o componente da IA no cerne da disrupção provocada na sociedade. Esse reconhecimento não apenas consagrou a significância da IA na era contemporânea, mas também destacou o papel crucial desses visionários em pavimentar o caminho para avanços que reverberam em todas as facetas da sociedade.

Este capítulo introduz conceitos básicos de IA (Seção~\ref{sec:fundamentos}) e, na Parte II, apresenta uma visão geral de suas implicações em diversas áreas: Visão Computacional (Seção~\ref{sec:visao}), Processamento de Linguagem Natural (Seção~\ref{sec:pln}), Saúde (Seção~\ref{sec:saude}), Indústria (Seção~\ref{sec:industria}), Finanças (Seção~\ref{sec:financas}) e Mobilidade Urbana (Seção~\ref{sec:transportes}). Usando aspectos da provocativa analogia de Ng, em cada área serão apresentadas algumas aplicações, riscos e tendências. 
A Parte III apresentam um panorama geral do trabalho, revisitando riscos em comum nas diferentes áreas; discute a IA neuro-simbólica como uma abordagem promissora pra esses riscos; e conclui com um chamado à reflexão sobre os rumos da tecnologia e da humanidade.
O capítulo não tem pretensões de esgotar o assunto, nem na listagem das áreas impactadas pela IA, pois praticamente todos os aspectos da vida em sociedade serão afetados, nem nas aplicações específicas de cada área. O conteúdo aqui apresentado é um convite para jornadas mais abrangentes e profundas do leitor, que poderá expandir seus horizontes nas referências apresentadas.

%\section{Fundamentos}  %nao descomentar aqui; o \section vem dentro do arquivo
\section{Fundamentos} 
\label{sec:fundamentos}

O psicólogo e economista Daniel Kahneman, ganhador de um Prêmio Nobel de Economia, propôs que o raciocínio humano é dividido em dois sistemas \cite{Kahneman2017thinking}. No Sistema 1, a mente trabalha de maneira instintiva, rápida, por reflexos, sujeita a erros e de maneira difícil de descrever (sem transparência). Este sistema é mais ativo em decisões rotineiras e tarefas mundanas, como os movimentos corretos a serem feitos enquanto se dirige um carro.
No Sistema 2, a mente trabalha de maneira deliberada, lenta, confiável e transparente.
Este sistema é mais ativo em decisões estratégicas e tarefas intelectuais, como a escolha de rotas enquanto se dirige um carro.

O modelo da mente humana dividida em Sistemas 1 e 2 é também útil para mapear abordagens de inteligência artificial \cite{geffner2018model}. 
Abordagens baseadas em aprendizado, mais modernas em IA, estão mais próximas ao Sistema 1: modelos treinados dão respostas rápidas, sujeitas a erros e difíceis de rastrear (baixa transparência).
Abordagens de IA simbólica, uma tradição dominante historicamente em IA até o início dos anos 2000, estão associadas ao Sistema 2: trata-se de métodos e algoritmos que enumeram explicitamente possíveis soluções para um problema e as investigam de maneira sistemática, sendo lentos, porém fáceis de rastrear (transparentes), pois é possível verificar o estado de um algoritmo e entender as decisões feitas por ele. Cabe ressaltar, no entanto, que Kahneman declarou que o Sistema 1 e Sistema 2 atuam de forma integrada, em debate sobre tendências em IA durante a conferência AAAI-2020 \footnote{AAAI-2020 Fireside Chat with Daniel Kahneman - com Francesca Rossi, Yoshua Bengio, Geoff Hinton e Yann LeCun. \url{https://vimeo.com/390814190} Acesso em 25 de setembro de 2023.}. Assim, embora exista a distinção entre Sistemas 1 e 2, os mesmos podem ser vistos de forma harmônica, o que nos remete à IA neuro-simbólica \cite{3rdWave}, que analisaremos na Seção \ref{sec:neurosimb}.

O restante desta seção apresenta fundamentos de IA simbólica (Seção \ref{sec:simbolica}) e IA baseada em aprendizado (Seção \ref{sec:ml}). Essa organização segue a ordem baseada na linha de tempo da IA, onde os métodos simbólicos (análogos ao Sistema 2 da mente humana) eram tradicionalmente predominantes, enquanto os métodos baseados em aprendizado (análogos ao Sistema 1 da mente humana) ganharam proeminência em tempos modernos, sendo responsáveis pela notoriedade atual da IA.
Ao leitor interessado em se aprofundar nos conceitos apresentados aqui, o abrangente livro de \cite{Russell&Norvig2020} é uma excelente referência para os conceitos de IA em geral.

%Os solucionadores são normalmente lentos, transparentes e gerais, enquanto as abordagens baseadas em aprendizado são normalmente rápidos, opacos e especializados. A integração sinérgica de solucionadores e abordagens baseadas em aprendizado poderia gerar sistemas inteligentes ainda mais fortes. Solucionadores são abordagens que aceitam modelos como SAT, programas lineares/inteiros mistos (MIP/IP/LPs) e planejamento automatizado, e produzem soluções automaticamente. Ou seja, solucionadores produzem soluções através de uma função mapeando entradas para saídas, e para isso requerem um modelo bem definido. Abordagens baseadas em aprendizado usam métodos de aprendizado de máquina para produzir funções com uma estrutura fixa que mapeiam entradas para saídas a partir de dados ou experiência. A modelo da mente humana dos Sistemas 1 e 2 é útil para compreender as relações entre solucionadores e abordagens baseadas em aprendizado \cite{geffner2018model}. Os solucionadores são normalmente lentos, transparentes e gerais, enquanto as abordagens baseadas em aprendizado são normalmente rápidos, opacos e especializados. A integração sinérgica de solucionadores e abordagens baseadas em aprendizado poderia gerar sistemas inteligentes ainda mais fortes.

%https://www-i6.informatik.rwth-aachen.de/~hector.geffner/www.dtic.upf.edu/~hgeffner/slides-invited-ijcai-2018.pdf
%Solvers and Learners
%
%\subsection{IA baseada em Raciocínio}
\subsection{IA simbólica}
\label{sec:simbolica}

\emph{Representação de conhecimento e raciocínio} é uma área da IA preocupada com como o conhecimento pode ser representado de forma simbólica e manipulado de maneira automatizada por programas que representam processos de raciocínio realizados sobre as representações simbólicas \cite{brachman2004knowledge}. Esta abordagem da IA busca estudar e produzir comportamento inteligente sem considerar necessariamente a estrutura biológica subjacente ao processo de raciocínio, mas focando no conhecimento que os agentes possuem. Assim, parte-se do pressuposto que o que permite aos agentes (como humanos) se comportarem de maneira inteligente é que eles sabem muitas coisas sobre o seu ambiente e são capazes de aplicar esse conhecimento conforme necessário para se adaptar às situações que se apresentam para alcançar seus objetivos. Portanto, nesta abordagem de IA, focamos no conhecimento e na representação deste conhecimento. As questões-chave nesta área são o que qualquer agente (humano, animal, artificial, etc) precisaria saber para se comportar de maneira inteligente e que tipos de mecanismos computacionais poderiam permitir que seu conhecimento fosse manipulado para realizar inferências que possibilitem que o agente atinja seus objetivos \cite{Fagin1995}.

Na área de representação de conhecimento e raciocínio, o conhecimento é geralmente visto como uma coleção de proposições, que são formulações  abstratas geralmente representadas por sentenças declarativas sobre o mundo (ou alguma parte ou aspecto dele) que podem ser verdadeiras ou falsas. Nesta área da IA, em geral, proposições são representadas por símbolos (como sequências de caracteres com sintaxe bem definida), que, ao contrário das proposições, são entidades concretas e que permitem a manipulação das proposições por meios computacionais \cite{Kowalski79}. Na abordagem simbólica, pesquisadores fazem utilização intensiva de formulações precisas, através das lógicas clássicas (proposicional e de predicados de primeira ordem) e lógicas não-clássicas, e.g. modais, temporais, epistêmicas, espaciais, probabilísticas, entre outras \cite{Broda2004}.
Assim, a representação de conhecimento é o campo de estudo focado em estudar o uso de símbolos formais para representar  proposições que constituem o conhecimento de um agente. Neste contexto, raciocínio é a manipulação computacional deste símbolos que representam o conhecimento de um agente, visando produzir novos símbolos, que representam um novo conhecimento. Uma coleção de representações simbólicas de conhecimento constitui o que chamamos de uma \emph{base de conhecimento}. Sistemas cujo comportamento inteligente é produzido pela manipulação das bases de conhecimento através de mecanismos computacionais que descrevem um processo de raciocínio são chamados de \emph{sistemas baseados em conhecimento}.  Neste tipo de sistema, adota-se uma abordagem declarativa, em que especificamos o que o agente sabe (onde fatos novos podem ser descobertos via percepção do ambiente) e o agente pode chegar a conclusões (que podem ser ações) através de inferências lógicas realizadas sobre sua base de conhecimento. Na abordagem declarativa de construção de sistemas, não especificamos o fluxo de controle da manipulação de dados, como em abordagens procedimentais.

Embora existam diversas abordagens, tipicamente a construção de sistemas ou agentes \emph{baseados em conhecimento} envolve o uso de linguagens formais baseadas em lógica (como a lógica de primeira ordem e lógica modal, entre outras) para representar conhecimento dependente de tarefa e domínio. Com isso, podemos definir mecanismos de raciocínio independentes de tarefa e domínio, que realizam processos computacionais que representam inferências lógicas bem fundamentadas que garantem propriedades lógicas desejáveis, como a validade. Estes mecanismos de raciocínio, por sua vez, são capazes de derivar novo conhecimento a partir dos fatos armazenados na base de conhecimento. Historicamente, a área de IA simbólica teve grande impulso a partir dos anos 1970 com desenvolvimentos em programação em lógica \cite{Kowalski79}. Especificamente, o desenvolvimento da linguagem Prolog (Programming in Logic) permitiu que pesquisadores passassem a expressar conhecimentos de domínios específicos de forma declarativa, sob uma fundamentação lógica e desenvolvessem sistemas computacionais a partir de uma abordagem simbólica. Prolog também teve grande impacto nos anos 1980 e 1990, notadamente durante o projeto liderado pelo Japão, denominado de "Fifth Generation Computer Systems Project". A linguagem Prolog foi aplicada com sucesso na prova automática de teoremas, sistemas especialistas, planejamento, bancos de dados, processamento de linguagem natural, aplicações legais entre outras áreas onde a representação de conhecimento e inferência lógica exigem uma representação computacional adequada \cite{Warren2023}.  

\emph{Planejamento automatizado} é uma subárea da IA simbólica que visa produzir solucionadores com comportamento direcionado a objetivos que sejam gerais e eficientes. Esses solucionadores, também chamados de planejadores, aceitam modelos que visam produzir comportamento direcionado a objetivos, sendo planejamento clássico um dos modelos mais pesquisados. Uma tarefa modelada por planejamento clássico possui um estado inicial, uma condição objetivo e um conjunto de operadores. Uma solução para uma tarefa de planejamento é uma sequência de operadores que satisfazem a condição objetivo quando aplicados ao estado inicial. Um exemplo de tarefa modelada no planejamento clássico é um cenário em um armazém onde o objetivo é encontrar uma sequência de movimentos que os robôs devem realizar para alcançar suas posições-objetivo.

A motivação do planejamento automatizado é criar um planejador que tenha um bom desempenho em qualquer tarefa sem conhecimento prévio \cite{hoffmann2011everything}. Isso torna o planejamento uma abordagem com bom custo-beneficio para desenvolvimento de soluções. Pode-se construir ou selecionar um planejador, descrever qualquer tarefa no modelo do planejador e então resolvê-la usando o planejador. Se a tarefa mudar, basta mudar o modelo da tarefa, mas não o planejador. Assim, utilizar planeadores para encontrar boas soluções para problemas do mundo real pode ajudar a reduzir tempo e custos.

Os planejadores mais bem-sucedidos baseiam-se em busca heurística. A$^*$ é o algoritmo de busca heurística mais conhecido~\cite{Hart1968}. Ele processa nodos de maneira sistemática, buscando o caminho ótimo (de menor custo) entre o estado inicial e o objetivo, empregando uma função heurística para descartar caminhos não-promissores. 
Heurísticas bem construídas aumentam a eficiência do algoritmo enquanto mantém sua otimalidade.
%Ele processa nodos em ordem de valor~$f$ não decrescente. A função~$f$ é a soma do custo atual do nodo raiz até o estado~$s$ contido no nodo~$n$ e o custo estimado para satisfazer a condição objetivo a partir do estado~$s$. Uma função heurística~$h$ é uma função que estima o custo para satisfazer a condição objetivo. A função heurística reduz o esforço necessário para encontrar uma solução, orientando o algoritmo de busca para processar primeiro os nodos mais promissores. Assim, a eficiência dos planejadores baseados em busca depende diretamente da qualidade da função heurística.

Funções heurísticas usam o modelo da tarefa para raciocinar automaticamente sobre a mesma, gerando estimativas do custo para satisfazer a condição objetivo. Em geral, este processo de raciocínio utiliza relaxações ou abstrações para calcular as estimativas em tempo polinomial~\cite{Helmert2009}. Por esta razão, os planejadores são chamados de independentes de domínio. Eles podem calcular estimativas diretamente do modelo sem conhecimento prévio sobre o domínio ou a tarefa.

%Falar de metodos classicos e.g. planning, representação de conhecimento, sist. especialistas, SAT, busca, CSPs, bayesian nets...

\subsection{IA baseada em aprendizado}
\label{sec:ml}

IA baseada em aprendizado é também conhecida como Aprendizado de Máquina. Em aprendizado de máquina, os sistemas computacionais são \emph{treinados} para realizar uma tarefa, ao invés de serem explicitamente programados para isso, como na IA simbólica.
O ponto chave em aprendizado de máquina é que os sistemas computacionais consigam realizar uma tarefa e melhorar seu desempenho nela à medida que adquirem mais dados ou experiência \cite{Mitchell1997}.
Esta seção apresenta uma visão geral da área, dividida nas três subáreas mais comuns: aprendizado supervisionado \ref{sec:supervised}, não-supervisionado \ref{sec:unsupervised} e por reforço \ref{sec:rl}. 

\subsubsection{Aprendizado supervisionado}
\label{sec:supervised}

Aprendizado supervisionado é uma área relacionada a tarefas de predição. Exemplos rotineiros incluem: predizer a probabilidade de chuva a partir dos dados climáticos, qual o valor de um ativo a partir de seus dados históricos, quais objetos estão presentes em uma imagem, qual a próxima palavra a se escrever a partir das palavras já escritas, entre outros.
Para se treinar um modelo de aprendizado supervisionado, é necessária a existência de um conjunto de dados rotulados, isto é, composto por instâncias definidas como pares de entrada e saída esperada. Através do seu processo de aprendizado, o algoritmo utilizará estes dados para encontrar padrões que mapeiam cada entrada para sua saída esperada.
Trata-se, portanto, de uma tarefa com \textit{feedback} instrutivo, onde, para cada entrada, há a instrução de qual é a saída ou resposta correta.

Quando a saída esperada no conjunto de dados faz parte de um conjunto finito de possibilidades, a tarefa de aprendizado supervisionado é de \emph{classificação}. Dentre os exemplos do início dessa seção, a predição de quais objetos estão presentes em uma imagem e qual a próxima palavra a se escrever a partir das palavras já escritas são tarefas de classificação.
Quando a saída esperada no conjunto de dados é um número, a tarefa de aprendizado supervisionado é de \emph{regressão}. Dentre os exemplos do início dessa seção, a predição da probabilidade de chuva a partir dos dados climáticos e do valor de um ativo a partir de seus dados históricos são tarefas de regressão.

Dentre as categorias de algoritmos de aprendizado supervisionado mais tradicionais, incluem-se (de maneira não-exaustiva):

\begin{itemize}
    \item Métodos baseados em instâncias, como ``k-vizinhos mais próximos'', onde não há um modelo treinado, mas um dado recebe a classe de seus vizinhos mais próximos;
    \item Árvores de decisão, onde o conjunto de dados é sucessivamente particionado de acordo com os valores do atributo escolhido em cada ponto (nó) de decisão;
    \item Métodos probabilísticos, como Naïve Bayes, onde a probabilidade de uma instância ser de uma dada classe depende das ocorrência de seus atributos em cada classe do conjunto de treino;
    \item Métodos de combinação de modelos (\textit{ensembles}), onde múltiplos modelos preditivos são combinados para melhor desempenho geral;
    \item Métodos conexionistas, como as redes neurais, que inspiram-se na capacidade de processamento de sinais do cérebro biológico.
\end{itemize}

A seguir, apresentamos uma breve descrição de redes neurais, visto que elas tem sido o componente principal dos sistemas mais modernos e disruptivos de IA. %, como o ChatGPT.

\subsubsection{Redes neurais e aprendizado profundo}
Redes neurais são modelos computacionais inspirados na estrutura e funcionamento do cérebro biológico. De maneira simplificada, os neurônios que compõem o cérebro biológico são elementos capazes de processar sinais elétricos, recebendo sinais de outros neurônios, atenuando-os ou amplificando-os e combinando-os em um sinal de saída a ser processado por outros neurônios.
\cite{McCulloch1943} foram pioneiros ao propor uma versão matemática desse elemento: o neurônio artificial recebe ``sinais'' numéricos como entrada, e combina-os em uma soma ponderada, na qual pesos numéricos fazem o papel de atenuar ou amplificar os números recebidos como entrada. 
Os pesos são os parâmetros do neurônio e podem ser modificados para melhorar seu desempenho.
Uma função de ativação é então aplicada a essa soma ponderada, resultando na saída do neurônio. 

Quando os neurônios artificiais são organizados em camadas interconectadas e a função de ativação é não-linear, temos o perceptron multicamada, o modelo mais tradicional de redes neurais. 
Quando o perceptron multicamada tem pelo menos uma camada intermediária (ou oculta) entre a entrada e a saída, já é possível dizer que trata-se de aprendizado profundo, pois tal rede já é capaz de detectar e combinar características não-lineares nos dados de entrada \cite{Goodfellow2016deeplearning}, embora alguns pesquisadores somente ``reconhecem'' como profundas as redes com dezenas de camadas.
%A Fig. \ref{fig:mlp} ilustra o perceptron multicamada.

%\todo[inline]{figura do MLP}

O treinamento de uma rede neural é uma forma de otimização baseada em gradiente. Nessa modalidade, uma função de custo, contínua e diferenciável, avalia as diferenças entre predições da rede e saídas esperadas. 
O gradiente dessa função de custo indica a mudança necessária em cada parâmetro (peso) da rede para que o custo, e consequentemente os erros da rede, diminuam.
Em termos práticos, o treino envolve a apresentação dos dados de entrada (números colocados na primeira camada), o cálculo das ativações da camada inicial até a final e a comparação com a saída esperada. 
Os pesos de todos os neurônios da rede são ajustados na direção determinada pela derivada parcial da função de custo com relação a cada peso. Essa derivada parcial é exatamente a mudança necessária em cada peso para que o custo diminua. Essas derivadas são calculadas da camada final da rede em retropropagação até a camada inicial. %A atualização dos pesos deve ser pequena e incremental, para o treino proceder de forma controlada. Dessa forma, normalmente o conjunto de dados é percorrido várias vezes para que se atinja um mínimo local da função de custo.
Esse procedimento é o clássico algoritmo \textit{backpropagation} de \cite{Rumelhart1986backprop}, cujos princípios são a base de praticamente todos os sistemas de aprendizado profundo. % que chocaram a humanidade em tempos recentes.

De uma maneira simplificada, uma rede neural é um conjunto de parâmetros (números) que realiza transformações numéricas em suas entradas. A clássica organização em perceptron multicamadas é uma das formas de se estruturar uma rede. Ela é especialmente eficaz para dados estruturados, com características já extraídas. %, até mesmo organizados de maneira tabular. 
Porém, para processamento de dados não-estruturados, a extração de características é necessária. 
Arquiteturas específicas de redes neurais são capazes de fazer essa extração de características, também chamada de aprendizado de representações. 
A seguir, descrevemos brevemente dois tipos de redes neurais que se tornaram modelos básicos em suas respectivas áreas: redes neurais convolucionais (CNNs do inglês \textit{convolutional neural networks}), extensivamente usadas em visão computacional (ver Seção \ref{sec:visao}) e Transformers, responsáveis por grandes avanços em processamento de linguagem natural (ver Seção \ref{sec:pln}).

%CLAUDIO THIAGO, por favor revisem abaixo
Em processamento de imagens, convoluções são operações matemáticas, nas quais filtros (ou \textit{kernels}) percorrem a imagem de entrada, na forma de uma janela deslizante, gerando um mapa de características. 
Os filtros definem uma característica de interesse a ser detectada e o mapa de característica resultante é uma ``imagem'', na qual os pixels da imagem original contendo a característica de interesse são destacados e os demais são atenuados. O objetivo é capturar padrões e características específicas, como bordas, texturas e formas, em diferentes partes da imagem. 
%O poder das convoluções reside na capacidade de compartilhar pesos e aprender características locais em vez de depender de um processamento global, uma vez que um filtro é capaz de detectar sua característica-alvo em qualquer lugar da imagem. 
\cite{LeCun1989cnn} foram pioneiros ao vislumbrar que os filtros convolucionais não precisavam ser pré-definidos, mas poderiam ser aprendidos via \textit{backpropagation} para extraírem características relevantes na tarefa em questão.
%Quando redes convolucionais são usadas em tarefas de classificação, normalmente um perceptron multicamada é usado após as camadas de convolução. Ele processa as características extraídas da imagem e as combina para realizar classificação. 
%A Fig. \ref{fig:cnn} ilustra uma rede convolucional em uma tarefa de classificação de imagens.
%Observe-se que após as convoluções, um perceptron multicamada é usado. Ele processa as características extraídas da imagem e as combina para determinar a classe da imagem em questão. 
Redes convolucionais estão no centro de várias aplicações em visão computacional, e o leitor interessado pode encontrar mais detalhes na Seção \ref{sec:visao}.

%VIVIANE por favor revise abaixo
Textos em linguagem natural são sequências de palavras, caracterizando-se como dados não-estruturados.
O aprendizado de representações em texto envolve uma série de desafios significativos, em grande parte devido à natureza complexa, variável e ambígua da linguagem natural.
%A tarefa de aprendizado de representações é especialmente desafiadora. 
Notáveis avanços foram feitos com a ideia de mapear uma palavra para um vetor cujas coordenadas no espaço dão uma ideia de significado \cite{Mikolov2013word2vec}. 
Nessa abordagem, palavras similares ficam em coordenadas próximas e há possibilidade de operações aritméticas entre palavras, como o clássico exemplo onde ``rei - homem + mulher = rainha''.
No entanto, essa abordagem tem a limitação de que palavras com diferentes significados são mapeadas para uma única representação, sendo insuficiente para saber, por exemplo se ``banco'' se refere à agência bancária ou ao local de se sentar.
Transformers \cite{vaswani2017attention} resolvem esse problema com o mecanismo de atenção: a representação de cada palavra não mais é fixa, agora ela depende do contexto (demais palavras anteriores e posteriores).
Tal mecanismo de atenção é composto de matrizes numéricas, cujos valores são aprendidos via \textit{backpropagation}. 
Transformers são o motor dos sistemas de processamento de linguagem natural mais impressionantes da atualizade, como o ChatGPT. A Seção \ref{sec:pln} apresenta mais informações sobre processamento de linguagem natural. % e o livro de \cite{Sutton&Barto2018} é a referência definitiva sobre aprendizado por reforço, para o leitor interessado.

%Em outras palavras, todos os sistemas de IA baseados em redes neurais são treinados com alguma variante do \textit{backpropagation} clássico.

\subsubsection{Aprendizado não-supervisionado}
\label{sec:unsupervised}

Aprendizado não-supervisionado lida com tarefas de descrição, em contraste com aprendizado supervisionado (predição) e por reforço (controle). 
Tarefas de descrição envolvem a extração de padrões, similaridades e informações ocultas nos dados sem a necessidade de rótulos ou supervisão explícita. De maneira sucinta, os principais tipos de tarefas descritivas são:

\begin{itemize}
    \item Agrupamento (Clustering): o objetivo é dividir o conjunto de dados em grupos de acordo com medidas de similaridade. Tipos de agrupamento incluem particional, no qual o espaço de estados é dividido em subregiões disjuntas, hierárquico, no qual a divisão particional pode ter múltiplas granularidades, e por densidade, na qual os dados são agrupados de acordo com a densidade (muita aglomeração ou dispersão). 
    \item Associação: o objetivo é identificar conjuntos de itens ou características que ocorrem juntos com alta frequência dentro de um conjunto de dados, revelando relações intrínsecas e estruturas subjacentes. Dentre as aplicações mais recorrentes destas técnicas, estão análise de cestas de compras e recomendação de produtos online. %Técnicas de associação são amplamente usadas em aplicações como análise de cestas de compras e recomendação de produtos online.
    \item Redução de Dimensionalidade: o objetivo é obter uma representação simplificada dos dados de entrada, reduzindo o número de dimensões (equivalente às ``colunas'' em conjuntos de dados tabulares) onde cada dimensão na nova representação é uma combinação das dimensões da representação original.
    Técnicas de redução de dimensionalidade são rotineiramente utilizadas para visualização e compressão de dados, sendo parte essencial do \textit{pipeline} de projetos em ciências de dados.
\end{itemize}

\subsubsection{Aprendizado por reforço}
\label{sec:rl}

Aprendizado por reforço (\rl) é geralmente associado a tarefas de controle, ou seja, aprender a melhor ação a se realizar em cada situação do ambiente.
Em contraste com aprendizado supervisionado, em \rl\ não se diz explicitamente ao aprendiz ou agente o que fazer ou qual a ação correta em uma dada situação. 
Ao invés disso, cada ação do agente é avaliada com um sinal numérico de \rew, o qual dá ideia de qualidade imediata daquela ação. 
%Tarefas de aprendizado por reforço envolvem, portanto \textit{feedback} avaliativo \cite{Sutton&Barto2018}.
O próprio \ag\ deve, por tentativa-e-erro, encontrar a ação que traz a maior quantidade de recompensas a longo prazo.

Dentre os métodos tradicionais de aprendizado por reforço, o clássico \ql\ \cite{Watkins&Dayan1992} é um dos pioneiros, e vários dos métodos mais bem sucedidos da atualidade usam seus princípios. 
No \ql, o agente mantém estimativas de valor (relacionado à soma das recompensas esperadas) para cada ação que possa executar em cada estado do ambiente. % $a$ quando o agente a realiza no estado $s$, denotado  $Q(s,a)$.
Ao interagir com o ambiente, o \ag\ obtém uma amostra da recompensa real da ação que realizou no estado que estava. O \ql\ usa essa recompensa, além do valor do estado atingido, para atualizar suas estimativas.
O \ql\ possui garantias teóricas de convergência de suas estimativas para os valores corretos \cite{Watkins1989}. 
Um agente treinado com \ql\ pode garantir o máximo possível de recompensa simplesmente selecionando a ação de maior valor em cada estado. 

O \ql\ mantém as estimativas de valor das ações para cada estado em uma tabela.
Se há muitos estados e/ou ações, tal representação não é viável.
Uma maneira de resolver isso é usar uma função ao invés de uma tabela para as estimativas de valor. Tais funções podem receber representações contendo características dos estados e aplicar pesos, que podem ser aprendidos, para ponderar essas características \cite[Parte II]{Sutton&Barto2018}.
Em especial, a própria representação do estado pode ser aprendida, por exemplo, por uma rede neural profunda. 
Essa abordagem é colocada em prática no algoritmo Deep Q-Networks (DQN) \cite{Mnih+2015}, o qual recebia pixels da tela e aprendeu a jogar jogos de Atari sem nenhum conhecimento prévio, obtendo desempenho sobrehumano em certos jogos. 

Pode-se dizer que DQN inaugurou a ``era do Aprendizado por Reforço Profundo'', onde avanços substanciais continuam acontecendo. Tais avanços levaram métodos de aprendizado por reforço a obterem grande sucesso em jogos, desde jogos de tabuleiro \cite{Silver+2017,Silver2017alphazero} até video-games muito mais complexos que Atari \cite{Berner2019dota,Vinyals+2019}, onde múltiplos jogadores devem responder rapidamente aos acontecimentos da tela enquanto traçam planos de longo prazo para vencer uma partida.

No entanto, a aplicação mais disruptiva de aprendizado por reforço foi em processamento de linguagem natural. 
Parte da metodologia de treino do ChatGPT consistiu em obter um modelo de recompensa através de \textit{feedback} humano para textos gerados por um modelo pré-treinado e o refinamento do modelo gerador de textos para maximizar esta recompensa \cite{Ouyang2022instructgpt}. 
%Essa metodologia é um componente importante do disruptivo em bem-sucedido ChatGPT, possivelmente o sistema de IA de maior impacto na história da humanidade.

Para o leitor intressado, o livro de \cite{Sutton&Barto2018} é o principal livro-texto sobre aprendizado por reforço.

\part*{Parte II: Impactos da IA}

Esta parte discute, de maneira não exaustiva, diversas áreas impactadas pela IA. Em cada área, são discutidas algumas aplicações, riscos e tendências. 
Inicialmente, discutimos ``áreas meio'', nas quais a IA tem relação com habilidades cognitivas humanas de visão (Seção \ref{sec:visao}) e linguagem (Seção \ref{sec:pln}).
Avanços nas referidas áreas têm reflexo nas ``áreas fim'', nas quais a IA afeta diferentes aspectos da sociedade: Saúde (Seção~\ref{sec:saude}), Indústria (Seção~\ref{sec:industria}),  Finanças (Seção~\ref{sec:financas}) e Mobilidade Urbana (Seção~\ref{sec:transportes}).

%\section{Visão computacional e Processamento de imagens} %nao descomentar aqui; o \section vem dentro do arquivo
\section{Visão computacional~e Processamento de imagens}
\label{sec:visao}

%Claudio e Thiago: revisar esse paragrafo introdutorio
Visão Computacional é um campo interdisciplinar que combina elementos de inteligência artificial, ótica e processamento de imagem. 
Trata-se de tornar as máquinas capazes de interpretar e extrair informações significativas a partir de imagens ou vídeos, permitindo a realização de tarefas como detecção de objetos, reconhecimento facial, análise de cenas, entre outras. 
%Este campo desafiador e dinâmico não apenas replica a habilidade humana de processar informações visuais, mas também estende suas aplicações a uma ampla variedade de setores, da medicina à robótica e à segurança, prometendo revolucionar nossa interação com o mundo digital e físico.

\subsection{Aplicações}

As áreas de visão computacional e processamento de imagens testemunharam enormes avanços nos últimos anos, em grande parte impulsionados por técnicas de aprendizado profundo. O aprendizado profundo, com sua capacidade de aprender automaticamente padrões complexos em grandes conjuntos de dados, revolucionou a maneira como abordamos as tarefas visuais. A sinergia entre visão computacional/processamento de imagens e aprendizado profundo levou a avanços significativos e abriu uma infinidade de aplicações práticas em vários setores e aplicações, algumas das quais brevemente listadas a seguir:
\begin{itemize}
\item Restauração e melhoria de imagens: algoritmos que integram processamento de imagens e aprendizado de máquina têm possibilitado a restauração visual de imagens degradadas, como remoção do ruído, aumento de resolução, correção de borramento por desfoco ou movimento, e correção de iluminação (sobretudo para imagens subexpostas).
%muito escuras).

\item Detecção, Segmentação e Reconhecimento de Objetos:
Algoritmos de detecção ou segmentação de objetos baseados em aprendizado profundo permitiram a identificação precisa e em tempo real de objetos em imagens e vídeos. Esta aplicação encontra uso prático em sistemas de vigilância, veículos autônomos e robótica. Por exemplo, carros autônomos usam a detecção de objetos para detectar pedestres, veículos e sinais de trânsito, e exploram segmentação para avaliar a área navegável. 

\item Reconhecimento facial e biometria:
O reconhecimento facial é amplamente utilizado para fins de identificação e autenticação. Modelos de aprendizado profundo, como redes neurais convolucionais (CNNs), são capazes de identificar atributos faciais discriminatórios de cada indivíduo,  permitindo o reconhecimento facial preciso mesmo em condições desafiadoras. Por exemplo, a biometria facial é empregada para autenticação de usuários em \textit{smartphones}, sistemas de segurança e aplicativos de controle de acesso, agilizando processos e aumentando a segurança.

\item Análise de Imagens Biomédicas: 
Conforme já discutido na Seção~\ref{sec:saude}, o aprendizado profundo fez contribuições significativas para a análise de imagens biomédicas, auxiliando os profissionais de saúde a diagnosticar doenças com mais precisão e eficiência. Os modelos de aprendizado profundo podem detectar anomalias em raios-X, ressonâncias magnéticas e tomografias computadorizadas, auxiliando na detecção precoce de condições como câncer, doenças cardiovasculares e distúrbios neurológicos. Além disso, a visão computacional também facilitou a análise das lâminas histopatológicas, ajudando os patologistas a identificar e classificar as células cancerígenas.

\item Realidade Aumentada (RA) e Realidade Virtual (RV): a visão computacional baseada em aprendizado profundo desempenha um papel crucial no desenvolvimento de aplicativos para RA e RV. Em particular, algoritmos de reconstrução tridimensional (3D) a partir de uma ou mais imagens podem ser usados para modelar ambientes reais, permitindo uma experiência imersiva com 
%capacetes
óculos
de VR. Além disso, permitem rastrear e reconhecer objetos e cenas do mundo real, possibilitando a inserção de objetos sintéticos no ambiente do usuário (RA).

\item Agronegócios e análise ambiental: a visão computacional baseada em aprendizado profundo transformou a agricultura com aplicativos como monitoramento de colheitas, detecção de doenças em vegetais e estimativa de rendimento. Drones equipados com câmeras e algoritmos de aprendizado profundo podem analisar vastas terras agrícolas, identificando áreas de preocupação e permitindo a aplicação precisa de fertilizantes e pesticidas. Além disso, o processamento de imagens aéreas e de satélite permitem o monitoramento de condições ambientais, como desmatamento, incêndios e enchentes.

\item Varejo e \textit{E-commerce}:
A visão computacional também encontra
%ra um
lugar nos setores de varejo e comércio eletrônico. Os varejistas podem usar a visão computacional para rastrear o comportamento do cliente em suas lojas, analisar o tráfego de pedestres e otimizar os \textit{layouts} das lojas para um melhor envolvimento do cliente. As plataformas de comércio eletrônico usam o reconhecimento de imagem para oferecer recomendações de produtos visualmente semelhantes, aprimorando a experiência de compra dos clientes.

\item Automação Industrial: técnicas de visão computacional podem ser utilizadas em ambientes industriais, para a identificação de defeitos em produtos de maneira rápida e eficiente. Além disso, conjuntamente com a robótica, por exemplo, pode-se fazer com que robôs equipados com câmeras acessem ambientes de difícil acesso e realizem monitoramento continuado da infraestrutura. 
\end{itemize}

\subsection{Riscos}

Apesar dos grandes avanços nos últimos anos, 
%há 
uma série de precauções
%que 
devem ser tomadas antes do uso irrestrito de algoritmos de visão computacional baseados em aprendizado de máquina.

Um potencial problema se refere a \textit{ataques}, nos quais uma imagem é manipulada para ``enganar'' um algoritmo de aprendizado de máquina. Por exemplo,  Eykholt e colegas~\cite{eykholt2018robust} apresentaram uma estratégia para realizar \textit{ataques físicos} focados no problema de detecção de sinais de trânsito (crucial para veículos autônomos). Como exemplo, mostraram que adesivos brancos e pretos colados a uma placa de trânsito mudam completamente o resultado de classificação de uma rede neural, apesar de manter o sinal completamente compreensível para o ser humano.

Um outro perigo potencial no uso de algoritmos de visão computacional baseados em aprendizado de máquina é a imprevisibilidade dos resultados em dados novos. Como a maioria das técnicas é supervisionada, são necessários dados de treinamento anotados. E como dados anotados são custosos de produzir e altamente dependentes do problema e aplicação, eles são disponibilizados em quantidade limitada. Por outro lado, aplicações de visão computacional envolvem dados nunca vistos pelas redes, que podem apresentar características distintas dos dados de treinamento, 
%que podem causar
potencialmente causando degradação da qualidade. Essa variabilidade de características dos \textit{datasets} é chamada de mudança de domínio (\textit{domain shift}), e pode envolver diversos parâmetros (e.g., treinamento em dados capturados durante o dia e teste com dados capturados durante a noite). Como exemplo, Hasan et al.~\cite{hasan2021generalizable} avaliaram o a capacidade de generalização de diversos algoritmos de detecção de pedestres, concluindo que a maioria apresenta resultados impressionantes em uma validação \textit{intra-dataset}, mas com degradação acentuada em validações \textit{cross-datasets}, mesmo quando o domínio alvo possui poucas diferenças visuais com relação ao domínio fonte.

Outra questão importante envolve os vieses, como os diversos problemas demonstrados por pesquisadores em sistemas de reconhecimento facial\footnote{"Study finds gender and skin-type bias in commercial artificial-intelligence systems:
Examination of facial-analysis software shows error rate of 0.8 percent for light-skinned men, 34.7 percent for dark-skinned women." MIT News, 11 de Fevereiro de 2018. {https://news.mit.edu/2018/study-finds-gender-skin-type-bias-artificial-intelligence-systems-0212}. Acesso em 28 de Agosto de 2023 }. Notadamente,  \cite{buolamwini2018gender} mostraram que algoritmos de reconhecimento facial discriminavam por raça e gênero. 

%\crj{Falta de previsibilidade, ataques, etc. }

\subsection{Tendências}

Por muito tempo,
a grande maioria das redes neurais envolvendo imagens era baseada em camadas convolucionais. Embora elas tenha uma representação compacta em termos de número de parâmetros e estejam relacionadas com o funcionamento do sistema visual humano, o uso de convoluções assume um filtro com suporte espacial limitado. Assim, \textit{pixels} muito distantes entre si podem não se relacionar em uma rede convolucional. Uma tendência crescente envolve o uso de camadas de atenção espacial, dentre as quais os \textit{Transformers} são populares. O modelo Visual Transformer (ViT)~\cite{dosovitskiy2020image} estende o conceito de 
\textit{Transformers}, originalmente desenvolvidos para textos, para o domínio de imagens.

Outra tendência atual é o uso integrado de dados visuais e textuais, dando origem às \textit{Vision Language Models}. Como exemplo, o modelo CLIP (\textit{Contrastive Language Image Pre-raining})~\cite{radford_learning_2021} usa uma base de 400 milhões de imagens pareadas com as respectivas descrições textuais, treinando os \textit{embeddings} de texto e de imagens de tal maneira que eles sejam similares. Com esse tipo de abordagem, se pode fazer consultas a imagens com dados de entrada textuais, e vice-versa. Com isso, diversos novos modelos e bases de dados têm sido propostas nos últimos anos. O \textit{dataset} LAION-5B, por exemplo, fornece 5,85 bilhões de pares texto-imagens. Em particular, se mostrou que redes profundas treinadas com uma quantidade muito grande de dados (como CLIP) podem ser customizadas com poucos dados adicionais para tarefas específicas. Tais redes são chamadas de \textit{Foundation Models}.

%\section{Processamento de Linguagem Natural} %nao descomentar aqui; o \section vem dentro do arquivo

\section{Processamento de Linguagem Natural}
\label{sec:pln}

Processamento de Linguagem Natural (PLN) é uma área multidisciplinar que combina ciência da computação e linguística com o objetivo de permitir que as máquinas realizem tarefas úteis com a linguagem humana. 
PLN geralmente envolve o processamento de grandes volumes de dados textuais ou de fala, que são chamados de \textit{corpus} (ou \textit{corpora} no plural). 
O PLN tem uma longa história e vem sendo estudado desde os anos 1950. 
A ideia de fazer com que os computadores consigam compreender a linguagem humana é tão antiga quanto a computação.
Alguns dos exemplos mais icônicos vieram da ficção científica, como o HAL-9000 do filme “2001: Uma Odisseia no Espaço” dirigido por Stanley Kubrick (1968) e escrito por Kubrick e Artur C. Clarke, baseado em obra anterior deste intitulada "The Sentinel". O computador HAL-9000 exibia uma série de habilidades impressionantes envolvendo a compreensão e a geração de linguagem. 
Nos últimos anos, algumas dessas habilidades foram enfim atingidas por algoritmos reais.
O imenso avanço em PLN se deve principalmente à evolução dos algoritmos de aprendizado profundo. Mais especificamente, o desenvolvimento da arquitetura de Transformers \cite{vaswani2017attention} usada nos grandes modelos de linguagem, do inglês \textit{large language model} (LLM), como BERT~\cite{devlin2018bert} e GPT~\cite{brown2020language}, melhoraram sensivelmente os resultados em diversas tarefas de compreensão e de geração de texto.
Pode-se dizer que as mudanças de maior impacto na imprensa e, notadamente disruptivas na computação nos últimos anos, vieram da área de PLN.

\subsection{Aplicações}
O PLN pode ser empregado em uma vasta gama de aplicações, tanto científicas como industriais.
A seguir, fornecemos uma lista não exaustiva de aplicações.

Diversas tarefas de PLN podem ser modeladas como problemas de \textbf{classificação de textos}, por exemplo: análise de sentimento, filtragem de spam, identificação de discurso de ódio, atribuição de autoria, detecção de plágio, \textit{etc}. Em todos esses casos, a entrada do algoritmo é uma sequência de tokens (\textit{i.e.,} palavras) e a saída é a classe predita. Até meados dos anos 2000, os algoritmos mais empregados para classificação de textos eram os de aprendizado de máquina tradicional como Naïve Bayes, máquinas de vetores de suporte e árvores de decisão. A partir de 2014, as redes neurais profundas como o LSTM \cite{hochreiter1997long} passaram a predominar. De 2019 para cá, a hegemonia de modelos baseados em Transformers, principalmente o BERT~\cite{devlin2018bert}, é notável. O ganho de qualidade obtido nessas tarefas pode ser atribuído à ideia de trabalhar em duas etapas: o \textit{pré-treinamento} e o \textit{ajuste fino}. No pré-treinamento, o algoritmo analisa um corpus de grande volume (\textit{i.e.,} contendo bilhões de palavras) a fim de aprender características da linguagem como relações entre palavras e coerência entre frases. O corpus usado nessa fase é apenas texto puro, que pode ser facilmente obtido a partir da Web. Uma vez que o modelo tenha sido pré-treinado com grandes corpora, ele pode ser ajustado para desempenhar uma tarefa específica, como análise de sentimento, por exemplo. Nessa fase, o modelo precisa de um conjunto de dados rotulado com a classe esperada -- o processo tradicional de aprendizado supervisionado. A vantagem aqui, é que o "conhecimento" que o modelo pré-treinado já possuía sobre a linguagem é aproveitado na tarefa específica e traz ganhos na qualidade das predições geradas.

A \textbf{sumarização de textos} tem por objetivo gerar uma versão reduzida do texto de entrada que contemple suas ideias principais. Diferentemente das tarefas de classificação em que a saída é uma só (\textit{e.g.}, a classe predita), na sumarização a saída, assim como a entrada, é uma sequência de tokens. Algoritmos de sumarização podem ser empregados em diversos domínios como financeiro, jurídico, científico para reduzir a quantidade de texto que as pessoas precisam processar.
As técnicas de sumarização dividem-se em \textit{extrativas} e \textit{abstrativas}. As técnicas extrativas usam técnicas estatísticas para selecionar as $k$ frases mais significativas do texto original e as usam para compor o resumo. A desvantagem é que a coerência do texto gerado é prejudicada pois a conexão entre as frases pode ser perdida. Já as técnicas abstrativas usam LLMs para tentar simular o comportamento de um sumarizador humano combinando sentenças e utilizando paráfrases para gerar um texto mais fluido. A desvantagem aqui é que esses modelos podem "alucinar" e adicionar conteúdo no resumo que não tem base no texto original (veja mais sobre esse problema na Seção~\ref{sec:riscos}).
    
A \textbf{tradução automática} (TA) é outra aplicação importante de PLN que obteve melhorias significativas com a introdução dos algoritmos de aprendizado profundo. A tarefa consiste em transformar uma sequência de um idioma fonte para um idioma alvo. O exemplo mais conhecido é de tradutor automático é o GoogleTranslate\footnote{\url{https://translate.google.com}}. Assim como a sumarização, tanto a entrada como a saída do algoritmo são sequências de tokens. 
A TA usa aprendizado supervisionado: durante o treinamento, o sistema processa um grande número de sentenças paralelas (as mesmas sentenças escritas no idioma fonte e no idioma alvo) e assim aprende o mapeamento entre idiomas. As arquiteturas mais usadas em sistemas de TA são LSTMs bidirecionais \cite{schuster1997bidirectional} e Transformers \cite{vaswani2017attention}. Além disso, vale ressaltar que esses sistemas trabalham em nível de subpalavras. Assim, eles conseguem aprender traduções para fragmentos, \textit{e.g.}, o sufixo de gerúndio "\textit{endo}" em português é normalmente traduzido para "\textit{ing}" em inglês.
A tradução automática é desafiadora por uma série de razões: ($i$) a ambiguidade que já é problemática em um idioma, fica ainda mais complexa quando adicionamos mais idiomas, ($ii$) a estrutura dos idiomas pode ser muito diferente e isso faz com que a ordem das palavras no idioma alvo possa ser muito diferente da ordem no idioma fonte, ($iii$) nem sempre há uma palavra correspondente no idioma alvo -- pode não haver nenhuma ou podem haver várias traduções possíveis e ($iv$) vocabulários de domínios específicos (\textit{e.g.}, medicina, computação, direito) dificilmente terão grandes volumes de sentenças paralelas para permitir a geração de mapeamentos precisos.

Os \textbf{chatbots}, também chamados de agentes conversacionais são programas que se comunicam com as pessoas usando linguagem natural. O exemplo mais marcante de chatbot é o ChatGPT\footnote{\url{https://chat.openai.com/}} que foi lançado em novembro de 2022 e em menos de dois meses já tinha mais de 100 milhões de usuários. O ChatGPT também é um modelo baseado em Transformers e suas capacidades vão além de conseguir manter diálogos, ele consegue escrever código de programas e compor músicas e poemas. 
Apesar de não ter representado um avanço científico (pois as tecnologias utilizadas já haviam sido justificadas), o impacto do ChatGPT foi gigantesco -- poucas ferramentas geraram tanto interesse como essa.
O mecanismo por trás dessas habilidades é a geração de texto que, assim como a geração de imagens, faz parte da IA generativa. A geração do texto usando LLMs é conhecida por  geração autorregressiva ou geração causal. Ela consiste basicamente em escolher a próxima palavra a ser gerada condicionada às escolhas anteriores e à pergunta feita pelo usuário. Esse processo é conhecido como \textit{next word prediction} (ou predição da próxima palavra). 
Essas escolhas dependem das estatísticas de ocorrência das palavras em grandes corpora de textos e em uma série de parâmetros que controlam a aleatoriedade do processo de geração.
Os riscos desses sistemas são discutidos na Seção~\ref{sec:riscos}.

Até agora, esta seção abordou apenas texto. Contudo, PLN também trata de fala. A habilidade de \textbf{reconhecer e produzir fala} são muito relevantes e úteis em uma série de aplicações que interagem com os usuários por meio de voz como assistentes inteligentes (como a Siri da Apple e a Alexa da Amazon).
A comunicação por meio de voz envolve o reconhecimento da fala (\textit{Automatic Speech Recognition}) (ASR) que transforma de fala para texto e a conversão de texto para fala \textit{Text to Speech} (TTS). ASR é uma tarefa bastante desafiadora pois precisa lidar com variações na forma de pronunciar as palavras (diferentes sotaques e velocidades de fala), ruídos de fundo e disfluências (sons como "hum" e "hã").
Tanto ASR como TTS atualmente são implementados utilizando LSTMs bidirecionais \cite{schuster1997bidirectional} ou Transformers \cite{vaswani2017attention}. Por ser mais difícil, ASR comumente precisa de mais dados de treinamento (\textit{i.e.,} mais horas de áudio pareadas com o texto correspondente).

\subsection{Riscos}\label{sec:riscos}
Os principais riscos associados ao uso de aplicações de PLN advém, principalmente, de quatro problemas: o viés dos dados, as alucinações, o potencial para mau uso e o custo do treinamento de LLMs.

Os LLMs são treinados com grandes volumes de textos coletados a partir da web. Esses dados não passam por um processo de curadoria e podem conter diversos tipos de \textbf{viés} (racismo, sexismo, homofobia, xenofobia, \textit{etc.}). O problema é que, ao gerar modelos a partir desses dados, os modelos passam a replicar esses vieses. 
% trecho comentado substituido pelo que vem logo abaixo: discussão do Prates et al expandida com o texto vindo da secao de fundamentos
%Por exemplo, \cite{prates2020assessing} observaram que o GoogleTranslate têm preferência por traduzir profissões relacionadas à ciência e tecnologia para o gênero masculino (\textit{e.g.,} engenheiro), mesmo quando não havia nenhuma especificação de gênero no idioma original.
%De forma similar, \cite{Papakyriakopoulos2020Bias} observaram que até mesmo representações geradas a partir de textos da Wikipedia apresentam sexismo, homofobia e xenofobia.
\cite{Papakyriakopoulos2020Bias} observaram que até mesmo representações geradas a partir de textos da Wikipedia apresentam sexismo, homofobia e xenofobia.
Também investigando vieses, mas na área de tradução automática, \cite{prates2020assessing} mostraram que o Google Translate apresentava uma forte tendência de tradução para \emph{defaults} masculinos em experimentos realizados a partir de uma lista abrangente de cargos do "Bureau of Labor Statistics" dos EUA. No artigo, traduções como “Ele/Ela é um Engenheiro” (onde “Engenheiro” é substituído por o cargo de interesse) em 12 idiomas diferentes de gênero neutro %, como húngaro, chinês e outros 
mostram que tradutor (que usa técnicas de IA) não consegue reproduzir uma distribuição real de trabalhadoras. %O artigo mostra por provas experimentais de que, Mesmo que não se espere, em princípio, uma distribuição pronominal de gênero de 50:50, 
O artigo mostra que o Google Translate produz padrões masculinos com muito mais frequência do que seria esperado apenas com base nos dados demográficos. 
Esses trabalhos indicam que é necessário o desenvolvimento de abordagens mais sofisticadas para a construção de sistemas que sigam princípios éticos, respeitando as diversidades populacionais, culturais, nacionais, de gênero, raça e muitas outras \cite{nature-ethics}. 

O segundo risco afeta os sistemas que geram texto de maneira autorregressiva: as \textbf{alucinações}.
As alucinações referem-se a situações em que um modelo de linguagem gera texto que contém informações que não estão presentes nos dados de treinamento, ou seja, o modelo gera fatos falsos. Há vários casos que foram divulgados na imprensa e mídias sociais envolvendo desde erros mais inofensivos até a imputação de crimes a pessoas inocentes.
A principal causa é a forma como esses modelos geram os textos: eles não têm nenhuma compreensão acerca da realidade que os textos descrevem.
Pesquisadoras críticas dessa abordagem referem-se a esses modelos como "papagaios estocásticos" \cite{bender2021dangers}.
É importante ressaltar que ferramentas que apenas geram texto não substituem motores de busca (como o Google e Bing, por exemplo) pois elas não têm como apontar as fontes para as informações. Quando solicitadas, elas podem até mesmo criar referências falsas.

A qualidade dos textos gerados automaticamente pode ser útil em uma série de tarefas, mas por outro lado, abre possibilidades para o mau uso.
Há relatos de advogados que usaram ferramentas como ChatGPT e Bard\footnote{\url{https://bard.google.com/}} para redigir processos, de candidatos a empregos que geraram currículos automaticamente contendo dados "inflados", de alunos que entregaram códigos de programa escritos pela ferramenta como sendo de sua autoria, de geração de notícias falsas, entre outros.

Por fim, com o aumento de ordens de grandeza no tamanho dos LLMs (\textit{e.g.,} de 117 milhões de parâmetros do GPT2 para 175 bilhões no GPT3 -- e um número desconhecido no GPT4) , o custo do treinamento desses modelos e o seu impacto ambiental também vêm sendo discutidos. Estimativas mencionam \cite{sharir2020cost} que o treinamento de um modelo com 1,5 bilhões de parâmetros possa chegar a US\$ 1,6 bilhões.

\subsection{Tendências}

Sob a perspectiva acadêmica, obter contribuições de impacto em PLN está cada vez mais difícil pois as universidades com seus orçamentos reduzidos precisam competir com gigantes do mercado de tecnologia como a Microsoft e Google. 
Levando isso em consideração, um artigo recente de pesquisadores da Universidade de Michigan \cite{ignat2023phd} aponta algumas futuras direções de pesquisa. Dentre elas, destacamos o desenvolvimento de modelos multilíngues e para idiomas com poucos recursos, a incorporação de um raciocínio que tenha fundamentação no mundo real para reduzir o problema das alucinações, o investimento em interpretabilidade dos modelos para possibilitar que as predições sejam explicadas e a aplicação de PLN em domínios relevantes como a saúde e a educação. 

A maturidade das técnicas de PLN e os bons resultados que vêm atingindo contribuem que elas sejam disseminadas e adotadas na indústria. A ampla disponibilidade de LLMs e modelos ajustados para as mais diversas tarefas facilita a sua implantação em sistemas, ferramentas e aplicativos que venham ser usado por um número cada vez maior de pessoas.

%\section{Saude}  %nao descomentar aqui; o \section vem dentro do arquivo
\section{Saúde}
\label{sec:saude}
A Saúde tem sido apontada desde cedo como uma das áreas de aplicação mais promissoras para a IA. Os primeiros exemplos de sucesso, ainda na década de 1970, tratavam-se de sistemas especialistas dependentes de conhecimento humano prévio e um conjunto de regras definidas para apoio à tomada de decisão clínica. Estes sistemas demonstraram utilidade para auxiliar na definição de diagnóstico ou na recomendação de tratamentos para pacientes, mas com um potencial muito limitado devido aos desafios de se representar um conhecimento complexo via regras e da incapacidade de extrapolar o conhecimento prévio a fim de aprimorar a tomada de decisão \cite{yu2018artificial}. 

Desde então, impulsionada pelo aumento na disponibilidade de dados em saúde e pelo rápido progresso de algoritmos capazes de aprender padrões relevantes e acionáveis a partir de dados volumosos e complexos, a IA vem gradualmente revolucionando a área da Saúde. Aplicações inovadoras baseadas em IA, especialmente em aprendizado de máquina, estão provocando mudanças significativas na forma como abordamos a medicina e os cuidados de saúde, sejam individuais ou coletivos. Esta seção revisa os aspectos principais da intersecção entre IA e Saúde no que tange aplicações, riscos e tendências.

\subsection{Aplicações}
%A Saúde tem sido apontada desde cedo como uma das áreas de aplicação mais promissoras para a IA. Os primeiros exemplos de sucesso, ainda na década de 1970, tratavam-se de sistemas especialistas dependentes de conhecimento humano prévio e um conjunto de regras definidas para apoio à tomada de decisão clínica. Estes sistemas demonstraram utilidade para auxiliar na definição de diagnóstico ou na recomendação de tratamentos para pacientes, mas com um potencial muito limitado devido aos desafios de se representar um conhecimento complexo via regras e da incapacidade de extrapolar o conhecimento prévio a fim de aprimorar a tomada de decisão \cite{yu2018artificial}. 

%Desde então, impulsionada pelo aumento na disponibilidade de dados em saúde e pelo rápido progresso de algoritmos capazes de aprender padrões relevantes e acionáveis a partir de dados volumosos e complexos, a IA vem gradualmente revolucionando a área da Saúde. Aplicações inovadoras baseadas em IA, especialmente em aprendizado de máquina, estão provocando mudanças significativas na forma como abordamos a medicina e os cuidados de saúde, sejam individuais ou coletivos. 

Embora praticamente todos os aspectos da prestação de cuidados de saúde sejam passíveis de uso e implementação de IA, quatro eixos se destacam nos esforços recentes, incluindo aqueles concentrados em países de baixa e média renda (LMICs, segundo sua sigla em inglês): (i) diagnóstico, (ii) avaliação do risco de morbidade ou mortalidade do paciente, (iii) previsão e vigilância de surtos de doenças e (iv) planejamento de políticas de saúde pública. \cite{schwalbe2020artificial}.

Sistemas para diagnóstico médico baseado em IA têm sido amplamente explorados em diversas áreas, mas alcançaram uma maturidade particular em especialidades como a radiologia, oftalmologia, patologia e dermatologia. Estes sistemas baseiam-se principalmente em dados de imagens médicas (e.g., ressonância magnética, tomografia computadorizada, fotografias de lesões ou de lâminas histopatológicas), demonstrando um desempenho diagnóstico via IA equivalente ao desempenho dos especialistas da saúde para casos de câncer de pele, câncer de mama, retinopatia diabética, doenças respiratórias, dentre outros \cite{liu2019comparison}. Sinais biomédicos (e.g., eletrocardiograma e eletroencefalograma), exames laboratoriais, dados genéticos (e.g., mutações no DNA e expressão gênica) e informações de prontuários médicos eletrônicos também foram utilizados com sucesso nas mais diversas especialidades médicas, e a evolução da IA vem possibilitando que os médicos façam diagnósticos mais rápidos e precisos. A IA foi empregada, por exemplo, para estratificação de risco em pacientes com infarto do miocárdio por oclusão \cite{al2023machine}, detecção precoce da doença de Alzheimer \cite{mahendran2022deep} e estimativa de risco de câncer de pulmão em 3 anos a partir de tomografia computadorizada e outras informações clínicas \cite{huang2019prediction}. 

A IA também tem sido uma tecnologia fundamental para aprimorar a capacidade de quantificar riscos de eventos desfavoráveis ou agravos relacionados à saúde de um paciente. Durante a pandemia da COVID-19, algoritmos de aprendizado de máquina foram amplamente aplicados para estimar quais pacientes infectados têm maior probabilidade de sofrer com uma doença mais severa ou vir a óbito pela COVID-19 ou suas complicações \cite{van2021artificial}. No estudo de \cite{phakhounthong2018predicting}, indicadores clínicos e laboratoriais foram utilizados para desenvolver um modelo baseado em IA para prever casos graves de dengue entre pacientes pediátricos durante a admissão. A avaliação de riscos propicia um melhor monitoramento do paciente e um tratamento mais efetivo através da antecipação de condutas clínicas, além de possibilitar uma melhor gestão de recursos hospitalares. 

Os benefícios da IA também podem ser observados na análise de riscos de Saúde em nível coletivo ou populacional, 
%Embora muitos dos exemplos mencionados sejam focados em uma avaliação individual de diagnóstico e prognóstico de doenças, a predição de riscos com IA também pode ser realizada em um nível coletivo ou populacional. 
%A IA pode contribuir 
contribuindo para o estudo da dinâmica de doenças e para uma melhor vigilância epidemiológica explorando uma grande variedade de dados, inclusive traços digitais (e.g., pesquisas na internet, atividades em redes sociais) \cite{brownstein2023advances}. \cite{jiang2018mapping} utilizaram aprendizado de máquina para estimar a probabilidade de surto epidêmico de Zika em nível global, conseguindo melhor modelar a complexidade e não-linearidade da relação entre o risco de transmissão por Zika vírus e fatores climáticos, ambientais e sócio-econômicos. \cite{brownstein2023advances} apontam que a IA tornou-se grande aliada na vigilância de doenças infecciosas, viabilizando o desenvolvimento de sistemas de alerta precoce para surtos de doenças, a identificação de focos de surtos ou de patógenos causadores de doenças, o rastreamento de contato e a previsão eficaz do risco de transmissão. Assim, a IA possibilita que autoridades de saúde pública respondam adequadamente ao risco que se apresenta, por exemplo, alocando recursos ou suprimentos adequados diante da expectativa de aumento de casos de uma determinada doença em uma região. Adicionalmente, o uso da IA possui grande impacto no planejamento de políticas de saúde pública, ao permitir a elaboração de medidas mais eficazes para proteção e promoção da saúde, como o planejamento de campanhas de vacinação e do direcionamento de materiais de divulgação de prevenção e cuidados com saúde com base no perfil de risco pessoal e padrões comportamentais \cite{panch2019artificial}.

Por fim, a IA se estende à análise de dados moleculares e genômicos, desempenhando um papel fundamental nas pesquisas biomédicas e possibilitando expandir nosso conhecimento sobre o funcionamento das doenças. A IA tem sido uma das principais propulsoras da medicina de precisão, especialmente na área da oncologia, revelando assinaturas moleculares associadas a subtipos ou estágios tumorais \cite{marczyk2023classification}, e identificando novos biomarcadores \cite{colombelli2022hybrid}, incluindo aqueles úteis para detecção não invasiva de câncer e avaliação de prognóstico \cite{xu2019translating}.
Adicionalmente, avanços recentes como o AlphaFold \cite{jumper2021highly}, que se utiliza de aprendizado profundo para prever com alta precisão as estruturas tridimensionais das proteínas a partir de sua sequência de aminoácidos, permitem facilmente avaliar o impacto funcional de variantes genéticas e acelerar a descoberta de novas drogas.  %Conhecer a estrutura de potenciais alvos é fundamental na descoberta de novas drogas, visto que a complementariedade química e de forma entre o alvo (usualmente uma proteína) e o ligante são determinantes para a afinidade da droga. 
%A IA também tem sido uma das principais propulsoras da medicina de precisão, especialmente na área da oncologia, revelando assinaturas moleculares associadas a subtipos ou estágios de tumores, e identificando biomarcadores que permitem detecção não invasiva de câncer e determinação do prognóstico de um paciente \cite{xu2019translating,rajpurkar2022ai}.

\subsection{Riscos}
Apesar do notável aumento nas pesquisas relacionadas às aplicações da IA na área da Saúde, é importante destacar que apenas um conjunto limitado destas soluções foi efetivamente implementado na prática clínica \cite{rajpurkar2022ai}. A Food and Drug Administration (FDA), agência reguladora vinculada ao Departamento de Saúde e Serviços Humanos dos Estados Unidos, tem desempenhado um papel ativo na revisão e autorização para comercialização de um número crescente de dispositivos médicos que incorporam IA. No entanto, até a última atualização em outubro de 2022\footnote{\url{https://www.fda.gov/medical-devices/software-medical-device-samd/artificial-intelligence-and-machine-learning-aiml-enabled-medical-devices}. Acesso em 14 de Agosto de 2023}, constatou-se a aprovação de apenas 521 dispositivos médicos pelo FDA. Dentre esses dispositivos, 56,23\% são voltados para aplicações em Radiologia e 10,94\% na Cardiologia. A disparidade evidente entre a extensa quantidade de pesquisas científicas conduzidas nesse domínio e o número limitado de soluções práticas adotadas reflete uma série de desafios que permeiam a integração da IA na prática médica. 

Embora diversos fatores possam contribuir para esse cenário, existe um consenso na comunidade acadêmica de que a falta de validação dos modelos baseados em IA por meio de dados externos constitui um dos principais fatores que inibem a aplicação efetiva do conhecimento científico adquirido \cite{liu2019comparison}. Esta validação deveria ser feita com dados prospectivamente coletados a partir do mundo real para este propósito específico, e seguindo uma metodologia criteriosa de avaliação, como aquelas adotadas em ensaios clínicos randomizados. Modelos de IA podem falhar na generalização para novos tipos de dados nos quais não foram treinados. Alguns trabalhos já demonstram que a capacidade preditiva de um modelo é impactada negativamente quando o modelo é aplicado a uma população de pacientes diferente dos seus dados de treinamento \cite{wong2021external}. Isto se deve à ampla heterogeneidade dos dados neste domínio devido a diferenças existentes nas práticas hospitalares e nos dados demográficos dos pacientes entre diferentes hospitais.  

Outro ponto crítico é que o treinamento de modelos de IA em conjuntos de dados com pouca representatividade de grupos marginalizados ou com variações injustificadas para determinados grupos resulta em sistemas tendenciosos que apresentam baixo desempenho preditivo nesses grupos. Assim, sem o controle adequado, o uso da IA introduz o risco de perpetuar vieses ocultos nos dados e reforçar preconceitos e desigualdades sociais existentes.
Por exemplo, um viés racial foi detectado em um algoritmo de avaliação de risco clínico utilizado nos Estados Unidos, que atribuía menor risco a pacientes negros em comparação com pacientes brancos igualmente doentes por utilizar custos de saúde como um proxy para as necessidades de saúde \cite{obermeyer2019dissecting}. Em outro estudo, um viés étnico foi identificado em escores de risco poligênico utilizados para estimar o risco de um indivíduo desenvolver doenças como câncer com base em fatores genéticos, possuindo acurácia muito superior em indivíduos de ascendência Europeia do que para outras ancestralidades em razão da coleta desequilibrada de dados genéticos e genômicos entre continentes \cite{martin2019clinical}. 

Estes riscos são exacerbados na impossibilidade de explicar a tomada de decisão pelos modelos e avaliar até que ponto a mesma reflete as abordagens humanas especializadas e não fere princípios éticos fundamentais. A Organização Mundial da Saúde (OMS) \cite{world2021ethics} chama atenção, ainda, para os vieses derivados de exclusão digital. Em alguns LMICs, mulheres têm menos acesso a telefone celular ou internet móvel do que homens, contribuindo com menos dados para treinamento de modelos de IA e sendo menos propensas a se beneficiar do uso desta tecnologia \cite{world2021ethics}.
A fim de gerar modelos baseados em IA que possam promover equidade em Saúde, é imprescindível garantir disponibilidade e qualidade de dados, com diversidade em relação a contextos sociais, culturais e econômicos. Por fim, é inevitável apontar o risco de violação da privacidade por se tratar de dados sensíveis, e o risco de uso indevido de dados pessoais, visto que muitas vezes os modelos são treinados com base de dados retrospectivas, originalmente coletadas para outros propósitos de pesquisa.

\subsection{Tendências}
A Organização Mundial da Saúde (OMS) \cite{world2021ethics} reconhece o enorme potencial da IA para promover melhorias na medicina e alavancar a equidade dos cuidados em saúde. Para que este potencial se concretize, avanços ainda se fazem necessários em diversas frentes, sendo algumas mais críticas para o domínio da saúde. 

Técnicas para mitigar vieses, sejam estes oriundos dos próprios dados ou resultantes do processo de treinamento dos modelos, são primordiais para evitar que se perpetuem desigualdades sociais existentes ou que se introduzam comportamentos tendenciosos nos modelos que possam produzir resultados discriminatórios contra determinados grupos. Entretanto, garantir maior equidade através do uso dos modelos também requer uma capacidade mais apurada de explicar as predições realizadas pelos mesmos a fim de identificar erros sistemáticos indesejáveis. Neste sentido, pesquisas em torno da explicabilidade de modelos são essenciais a fim de expandir a capacidade de encontrar fatores relevantes para as predições realizadas com base não somente em associações, mas em relações causais entre as variáveis de entrada e o resultado do modelo. Uma explicabilidade baseada em causalidade tornaria a interação especialistas-IA muito mais efetiva para a investigação da tomada de decisão feita pelos modelos, resultando em maior confiança na implementação prática destes modelos. 

Por fim, salienta-se que a tomada de decisão em um ambiente clínico é inerentemente baseada em múltiplas evidências, sendo portanto crucial ampliar a capacidade dos algoritmos de aprenderem a partir de dados multimodais. Desta forma, modelos multimodais de IA visam possibilitar o uso de todas as fontes de dados normalmente disponíveis aos médicos ou que possam enriquecer a definição de um diagnostico, como dados clínicos e laboratoriais, exames de imagens, testes genéticos, determinantes sociais, fatores ambientais ou comportamentais, informações coletadas por \textit{wearables}, dentre outros. Estas direções de pesquisa estão entre os principais pontos de acesso para fornecer cuidados ao paciente mais oportunos, precisos e justos com auxílio da IA.

%\section{Jogos} %nao descomentar aqui; o \section vem dentro do arquivo
%\input{sections/jogos}

%\section{Indústria} %nao descomentar aqui; o \section vem dentro do arquivo
\section{Indústria}
\label{sec:industria}

Da mesma forma que a eletricidade transformou drasticamente a indústria na segunda revolução industrial, a IA tem sido apontada como uma das promotoras de grandes transformações na indústria atualmente. Nos últimos anos, temos testemunhado uma crescente adoção de abordagens de IA nos mais diferentes setores da indústria, tais como energia \cite{pivetta2023, rahmanifard2019application}, manufatura \cite{li2017applications}, %, wuest2016machine}, 
indústria química \cite{baum2021artificial}, agricultura industrial \cite{benos2021machine}
, etc. 

Atualmente, a IA é apontada como um fator viabilizador com papel crucial na chamada \emph{indústria 4.0} (quarta revolução industrial), que tem como característica fundamental o foco no desenvolvimento de \emph{indústrias inteligentes}. Este cenário surge graças ao desenvolvimento e integração da IA com diferentes tecnologias, tais como redes de dados, internet das coisas, computação em nuvem, automação de processos físicos, sistemas ciber-físicos, etc. 

Em um cenário típico de uma indústria inteligente alinhada à industria 4.0,  a fábrica é constituída por coleções de sistemas ciber-físicos, que estabelecem uma interação profunda entre processos físicos e processos computacionais. Neste contexto, processos computacionais distribuídos controlam elementos físicos e sensores realizam continuamente o monitoramento dos processos e componentes físicos, retroalimentando os processos computacionais. Neste contexto, a IA desempenha um papel crucial na tomada de decisão que controla estes processos continuamente \cite{Anjos-ACM-092023}.

A seguir, serão discutidas algumas aplicações de IA na indústria, bem como os riscos e as tendências associadas ao uso de IA neste contexto.

\subsection{Aplicações}

A IA vem sendo aplicada nos mais diversos setores industriais, das mais diferentes formas, incluindo otimização e automação de processos produtivos, previsão de demandas e de produção, identificação de perfil de clientes, desenvolvimento de produtos com IA, automação de atendimento ao cliente, desenvolvimento de novos produtos, etc. Nestes contextos, a aplicação de técnicas de IA visa aumentar a produtividade, reduzir custos, tornar as linhas de produção mais seguras, etc.

Uma das aplicações mais notórias da IA na indústria diz respeito ao uso destas tecnologias para automação de processos produtivos \cite{ribeiro2021robotic,fragapane2022increasing}. A automação, neste caso, envolve principalmente a utilização de robôs, ou sensores e atuadores distribuídos ao longo das linhas de produção. Neste cenário, sistemas de IA utilizam dados de sensores para tomar decisões e controlar os atuadores ao longo da linha de produção.  

Técnicas de IA também vêm sendo largamente utilizadas na indústria para detectar anomalias em comportamentos de sistemas, de processos produtivos, etc \cite{stojanovic2016big, zipfel2023anomaly}. Anomalias, neste cenários, são padrões de comportamento diferentes do comportamento esperado \cite{marti2015anomaly}. Nestes contextos, em geral, são aplicadas técnicas de aprendizado de máquina para treinar algoritmos que identifiquem estados normais e anômalos dos sistemas e processos de interesse. Estes algoritmos costumam ser treinados a partir de dados de sensores que caracterizam os estados dos processos e sistemas ao longo do tempo. Esta abordagem é utilizada, por exemplo, para detecção em tempo real de possíveis vazamentos em oleodutos \cite{aljameel2022anomaly}. Em alguns casos, anomalias podem ser detectadas do modo visual também \cite{roth2022towards}, em cenários em que as anomalias não são bem representadas por medidas de sensores convencionais (como medidas de pressão e temperatura, etc), mas se tornam aparentes através da inspeção visual. Estas abordagens são muito comuns, por exemplo, para detectar defeitos em produtos em linhas de produção \cite{birlutiu2017defect}, permitindo a remoção do produto defeituoso do processo para eventuais correções dos defeitos. Nestas abordagens, técnicas de aprendizado de máquina podem ser utilizadas para aprender padrões que caracterizam produtos com e sem defeitos a partir de grandes conjuntos de imagens previamente rotuladas.

Além de aperfeiçoar os processos produtivos, tecnologias de IA também vêm sendo utilizadas na indústria para a previsão de demandas e para o gerenciamento da cadeia de suprimentos necessários para suprir estas demandas \cite{zhu2021demand, toorajipour2021artificial}, incluindo a previsão de oferta de insumos e seleção de fornecedores. Muitas das aplicações nesta área vêm utilizando técnicas de aprendizado de máquina capazes de lidar com dados em séries temporais. 

No contexto da indústria 4.0, os sistemas produtivos tendem a ser altamente sensorizados, de modo que uma grande quantidade de medidas são continuamente adquiridas dos equipamentos ao longo do tempo. Neste cenário, estes dados podem fornecer valiosos \emph{insights} sobre o estado dos equipamentos. Estes fatores vêm permitindo o desenvolvimento de técnicas de \emph{manutenção preditiva} \cite{paolanti2018machine,paolanti2018machine,serradilla2022deep,dalzochio2020machine} baseadas em técnicas de aprendizado de máquina. Abordagens de manutenção preditiva visam monitorar o estado dos equipamentos com o intuito de prever eventuais momentos de falha antes que elas ocorram, permitindo a redução de custos oriundos de paradas não programadas na produção, ou ainda evitando falhas que podem comprometer drasticamente as plantas de produção.

Nos últimos anos, a IA vem sendo utilizada até mesmo no processo de  \emph{design} de novos produtos \cite{aphirakmethawong2022overview}. Aplicações típicas de IA em design de produtos vêm utilizando as mais diversas abordagens de IA, incluindo desde algoritmos genéticos \cite{kielarova2023genetic} a aprendizado de máquina \cite{zhang2019food, fournier2021machine, hamolia2021survey}. Cabe destacar que nos últimos anos técnicas de IA generativa vêm demonstrando capacidades impressionantes em tarefas de design \cite{grisoni2021combining}. Modelos de IA generativa, como ChatGPT e Dall-E, são capazes de aprender padrões a partir de grandes massas de dados (imagens, textos, etc) e gerar saídas que reproduzem esses padrões de forma verossímil. Algumas aplicações representativas de IA em design de produtos incluem o projeto de circuitos integrados \cite{gubbi2022survey,wang2019machine,hamolia2021survey}, desenvolvimento de novas drogas \cite{grisoni2021combining}, desenvolvimento de peças de vestuário na indústria da moda \cite{liang2020implementation,giri2019detailed}, desenvolvimento de produtos na indústria alimentícia \cite{zhang2019food}, etc.

É importante salientar que as aplicações industriais de técnicas de IA são vastas, abrangendo muitos setores e muitas tarefas diferentes, de modo que nesta seção são discutidos apenas alguns exemplos.

\subsection{Riscos}
%https://aiindex.stanford.edu/wp-content/uploads/2023/04/HAI_AI-Index-Report_2023.pdf
%page 207
%Uma boa parte dos riscos que eu ia citar aqui já estão citados em outros:

A aplicação da IA na indústria herda boa parte dos riscos da IA aplicada em contextos gerais. Um destes riscos, e que pode impactar aplicações industriais de diversas formas, é o da falta de generalização de modelos de aprendizado de máquina. Em caso de falha na generalização destes modelos, sistemas de IA podem cometer erros em casos em que precisam lidar com situações muito diferentes das representadas nos dados de treinamento ou com dados capturados por sensores com características técnica diferentes dos sensores que coletaram os dados de treinamento. Em contextos industriais, erros no processo de decisão acarretados por modelos que não generalizaram adequadamente podem causar diversos impactos negativos. Por exemplo, em casos em que o ambiente industrial possui atuadores controlados por modelos sem a devida generalização, falhas no processo de decisão podem disparar ações (como movimentos de braços robóticos) que podem eventualmente ferir seres humanos que também atuam no ambiente industrial \cite{franklin2020collaborative}. Outros exemplos do impacto negativo da falta de generalização incluem detectar incorretamente defeitos em produtos, o que pode fazer com que produtos defeituosos sejam mantidos ou produtos sem defeitos sejam removidos das linhas de produção. 

Apesar da extensa pesquisa na área de aprendizado de máquina visando encontrar maneiras de mitigar o problema da generalização, ainda existem diversos desafios relacionados à própria identificação adequada do domínio de validade dos modelos de aprendizado de máquina. Ou seja, dado um modelo de aprendizado de máquina treinado em um certo conjunto de dados, não é trivial determinar quais são os conjuntos de situações em que ele funciona adequadamente ou não. Esta dificuldade pode tornar modelos de aprendizado de máquina suscetíveis aos chamados ataques adversários \cite{narodytska2017simple}, em que alguém mal intencionado pode alterar sutilmente os dados de entrada (de um modo imperceptível para seres humanos) de certos modelos com o intuito de perverter o comportamento esperado. Estas dificuldades estão em grande parte associadas à dificuldade de se explicar de forma significativa o que de fato foi aprendido pelo modelo. Essa dificuldade associada à explicabilidade de modelos de aprendizado de máquina vem sendo apontada como um risco pela indústria em geral.

Além disso, atualmente há uma grande discussão a respeito das consequências da aplicação da IA no mercado de trabalho \cite{agrawal2019artificial}. A utilização da IA na indústria, em geral, promove um aumento da automatização dos mais diferentes processos. No passado, os processos de automatização atingiram principalmente os aspectos físicos dos processos industriais. Mas com aplicações de tecnologias de IA estamos testemunhando também a automação (ou pelo menos uma aceleração) de aspectos intelectuais. Esta tendência gera ainda mais impactos na oferta de empregos, diminuindo a oferta de certos postos, mas eventualmente proporcionando o surgimento de novas profissões.

\subsection{Tendências}
%%Begin Lamb escreveu
Uma pergunta-chave, tendo em vista a ubiquidade da IA, é como identificar  tendências relevantes para negócios? Um  análise ampla de tendências na área de IA pode ser realizada de diversas formas. Muitas vezes, a abordagem acadêmica utilizada é da identificação de áreas de classificação de artigos em publicações. No entanto, esta é uma abordagem obviamente limitada às bases de consulta e a preferências das conferências e revistas no que se refere a áreas de pesquisas. 
O atual impacto da IA - ressalte-se - surge a partir de uma subárea que era pouco valorizada na academia por um período de mais de uma década: redes neurais artificiais. Entre meados da década de 1990 até 2006, quando Geoffrey Hinton e seus alunos publicaram o primeiro artigo no qual os autores se referem a redes neurais profundas\footnote{Deep neural networks, no caso do artigo \cite{Hinton_2006} se referem a "deep belief networks", uma arquitetura de redes neurais para aprendizado.} \cite{Hinton_2006}, poucos autores consideravam o aprendizado conexionista como sendo uma grande tendência futura em IA. Assim, pensar em tendências, como diria Niels Bohr, é prever o futuro - e não há algo mais difícil do que prever o futuro em ciência.\footnote{"Prediction is very difficult, especially if it’s about the future." Frase atribuída a Niels Bohr e, também, a Yogi Berra.}

Consultorias especializadas em tecnologia, como Gartner, IDC, McKinsey e diversas outras analisam e identificam periodicamente diversas áreas da computação que terão impacto ao longo do tempo, bem como sua maturidade \footnote{\url{https://www.gartner.com} e \url{https://www.idc.com/}}. Do ponto de vista de mercado, tais estudos têm grande relevância, pois orientam profissionais e gestores. Na última década, outros relatórios sobre análise e tendências em IA têm sido publicados por centros de pesquisa, reunindo parcerias entre a academia e as empresas. Entre eles destacamos o AI Index Report, produzido sob a coordenação do Human-Centered AI Institute da Universidade de Stanford\footnote{\url{https://aiindex.stanford.edu/report/}}. Este relatório, publicado anualmente desde 2017, destaca as tendências em pesquisa e desenvolvimente (através de análise de publicações), performance técnica (onde se analisam os progressos tecnológicos e seus impactos), ética em IA (equidade, viéses e suas implicações), impacto econômico (utlização da IA em negócios, investimentos públicos e privados), educação (nas escolas e universidades), políticas e governança (estratégias nacionais e multilaterais de governos), diversidade (notadamente na academia e as iniciativas para seu incremento) e opinião pública (análise da percepção pública sobre o impacto da IA). Uma observação relevante do AI Index Report\footnote{AI Index Report 2023, Capítulo 1, Página 50. \url{https://aiindex.stanford.edu/wp-content/uploads/2023/04/HAI_AI-Index-Report_2023.pdf}} é que até 2014, a maior parte dos sistemas de aprendizado de máquina eram produzidos pela academia. Desde então, as empresas passaram a dominar a produção destas tecnologias. Os dados do relatório indicam que produzir sistemas de aprendizado de máquina de estado-da-arte requer grandes volumes de dados, poder computacional e recursos financeiros não disponíveis às universidades.
Isto pode sugerir que, assim como demais tecnologias do passado, a partir do momento em que a viabilidade técnica e o alto potencial econômico de uma tecnologia são demonstradas, os investimentos neste setor tendem e se consolidar nas empresas. 
%%End Lamb escreveu

%\section{Finanças} %nao descomentar aqui; o \section vem dentro do arquivo
\section{Mercado de Capitais e Finanças}
\label{sec:financas}

Finanças tem dois aspectos interessantes, sendo ao mesmo tempo uma arte e uma ciência. Então, pode-se entender finanças como a arte e a ciência da gestão de ativos financeiros. Mas as pessoas frequentemente se deparam a questão recorrente a seguir: \textit{porque eu deveria me interessar pelo que acontece no mercado de capitais?}. Para abordar esta questão seria interessante discutir o que acontece no mercado de capitais.

O mercado de capitais é para onde os governos recorrem para fechar suas contas (geralmente pedindo empréstimos pela venda de pequenos ‘pedaços’ da dívida do governo, ex: títulos de dívida como os conhecidos ‘títulos do tesouro’). A razão principal para o governo ir ao mercado de capitais é honrar seus compromissos, tais como pagar benefícios sociais (ex: aposentadorias, seguro-desemprego, etc.), pagar outras contas obrigatórias (ex: saúde, educação, etc.), ou desenvolver seus projetos (ex: casa própria, saneamento básico, etc.). As empresas públicas e privadas também recorrem ao mercado de capitais para pedir empréstimos e cumprir com suas obrigações (ex: pagar dívidas vencendo em prazos curtos), desenvolver seus projetos (ex: investir em serviços licitados tais a expansão da rede de saneamento básico, do sistema de geração e distribuição de energia, etc.). As pessoas físicas também recorrem ao mercado de capitais para obter recursos e atingir diversos objetivos (ex: projetos futuros, casa própria, aposentadoria, etc.).

Como o mercado de capitais mobiliza a poupança, gere riscos, aloca eficientemente  recursos e promove o aumento da disciplina corporativa, toda a sociedade é beneficiada. Ao aplicar sua poupança em capital produtivo, os investidores (individuais ou institucionais) causam movimentos de capitais e buscam uma alocação eficiente e com menor custo. Isso aumenta a liquidez da economia e os prazos dos investimentos. Para fazer essa alocação de capitais (recursos), os participantes do mercado exigem qualidade na governança corporativa e o compartilhamento de informação por parte das empresas que captam estes recursos, levando a maior disciplina e transparência, o que impacta na produtividade e no retorno sobre o investimento realizado. O resultado final, em um nível macro-econômico, é mais emprego, renda, investimento e crescimento econômico, o que impacta positivamente os principais indicadores socioeconômicos do país.
Esse ciclo virtuoso apontado acima também traz resultados indiretos, tais como a viabilização e o desenvolvimento de mais projetos, a expansão da produção e da criação de riqueza, a criação de mais empregos, e o aumento da renda do cidadão. Portanto, desenvolvendo o mercado de capitais, promove-se o desenvolvimento socioeconômicos do país.

Então, parece natural que alguém tenha interesse pelo que acontece no mercado de capitais e no mundo das finanças, mas para identificar a relação entre a computação, o mercado de capitais e o mundo das finanças, precisamos observar mais detalhadamente  o que acontece no dia-a-dia do mercado de capitais.

Qualquer transação no mercado de capitais envolve um acordo entre duas partes: (a)  tomador de capital e (b) provedor de capital. Geralmente, quem provê capital o faz esperando algum retorno (ex: um valor referente ao aluguel do capital via juros, um lucro para remunerar o capital investido em uma transação, etc.). Por outro lado, quem toma capital também o faz esperando algum retorno (ex: atingir o objetivo de adquirir um bem de capital ou ativo real, desenvolver um projeto, etc.). Como há muitas partes interagindo direta ou indiretamente no mercado de capitais, com propósitos muito diferentes, existe uma grande variedade de instrumentos disponíveis para atender aos diferentes interesses das partes, possibilitado que elas interajam e atinjam seus objetivos através do uso de instrumentos específicos. Estas interações podem ocorrer em diferentes ambientes (ex: há ambientes em que existe livre negociação de instrumentos entre partes, como nas bolsas de valores, e existem ambientes onde a negociação ocorre dentro de restrições, como no caso das interações entre as empresas e seus clientes).  Esta diversidade de tipos de instrumentos transacionados entre partes e o grande volume de transações tendem a tornar o dia-a-dia do mercado de capitais complexa. Cada instrumento transacionado entre partes provê informações sobre o mercado, seus segmentos, partes envolvidas e sobre a própria economia, em diferentes níveis (ex: micro ou macro-econômico, local, regional, nacional ou mesmo internacional). Devido a grande complexidade das operações realizadas, da necessidade de rastrear e armazenar o enorme volume de informações gerado, o mercado de capitais migrou quase em sua totalidade para o ambiente digital e as transações são computadorizadas.

\subsection{Aplicações}

Hoje, a IA participa da automação de tarefas rotineiras, prove acessibilidade a serviços, e capacidade de aprendizado para aperfeiçoar tarefas rotineiras nos mercados, de forma exata, eficiente e rápida. Alguns dos temas abordados com o auxílio da IA no dia-a-dia do mercado de capitais e em finanças são: 1) análise ou inferência de sentimento dos participantes do mercado com base em textos, ou mesmo de mídias sociais; 2) detecção de transações suspeitas ou fraudulentas, ameaças e/ou crimes financeiros, ou ainda ameaças cibernéticas; 3) identificação de riscos e vantagens potenciais de ativos transacionados nos mercados; 4) avaliação e recomendação de instrumentos financeiros (ex: produtos e serviços) para potenciais interessados(as), com base nas suas preferências e objetivos; 5) processamento de dados estruturados e não estruturados, tais como documentos, para extrair dados relevantes e alimentar os processos de análise, previsão e recomendação (ex: descoberta de oportunidades de investimento); 6) uso de  estimativas de risco, dados de transações e complementares para prever com razoabilidade resultados futuros; 7) sintetize de informações relevantes para a tomada de decisão usando IA generativa.

Por exemplo, já é comum alguém ‘falar’ com um robot ao tentar abrir uma conta bancária pela internet, ou mesmo ser entrevistado(a) por um robot para direcionar uma chamada telefônica a uma instituição financeira. O Business Insider\footnote{\url{https://www.insiderintelligence.com/insights/ai-in-finance/}} estima que as aplicações de IA vão economizar para as instituições financeiras nos EUA, em 2023, cerca de USD 447 bilhões. A maioria dos bancos (cerca de $80\%$) avaliam como positivo o impacto da IA neste segmento da indústria, e pretendem acelerar a migração das suas operações físicas para o ambiente digital (ex: um dos maiores investimentos do Banco Mercantil do Brasil em 2023 é focado no aumento da acessibilidade digital). Mani Nagasundaram (Senior VP, Global Financial Services, HCL Technologies) afirmou recentemente em um artigo na AI News\footnote{\url{https://www.artificialintelligence-news.com/2020/12/15/from-experimentation-to-implementation-how-ai-is-proving-its-worth-in-financial-services/}} que a IA tende a liberar pessoal de tarefas rotineiras, e ao mesmo tempo melhorar a qualidade e a segurança do acesso a serviços financeiros, além de contribuir para trazer inovação ao ambiente corporativo. Já a Forbes\footnote{\url{https://www.forbes.com/sites/louiscolumbus/2020/10/31/the-state-of-ai-adoption-in-financial-services/?sh=711a8c4e2aac}} sugere em um artigo recente que $70\%$ das empresas do setor financeiro já usam IA para prever eventos que possam afetar o seu fluxo de caixa (antecipando a necessidade de caixa para as operações diárias), ajustar os limites de crédito dos clientes e detectar fraudes no uso dos serviços (ex: cartões de crédito).  Também, de acordo com a Forbes\footnote{\url{https://www.forbes.com/sites/jayadkisson/2019/01/23/artificial-intelligence-will-replace-your-financial-adviser-and-thats-a-good-thing/?sh=1a02118e6b40}}, IA tem sido usada para identificar as tendências mais recentes dos mercados, e avaliar os perfis das carteiras de investimento disponíveis e dos clientes, para então sugerir os quais  investimentos seriam adequados para cada perfil de cliente. Como IA é usada frequentemente para analisar padrões em grandes conjuntos de dados, naturalmente esta habilidade da IA tem sido usada em negociações nos mercados abertos (ex: bolsas). Pois os métodos computacionais baseados em IA podem analisar dados mais rápido e com maior exatidão que os humanos, além de poderem aprender a serem mais eficientes e otimizar as negociações nestes mercados (ex: algoritmos inteligentes já são usados para achar interessados em fixar a taxa de conversão Dolar-Real de um contrato de exportação, e os interessados em contratar esta taxa de conversão Dolar-Real na data futura desejada; ou ainda, algoritmos inteligentes já são usados para identificar tendências nos mercados e sugerir transações de ativos financeiros que estejam alinhadas com estas tendências). 

\subsection{Riscos}

Não se pode ignorar que existem desafios éticos e riscos a serem mitigados para que o uso da IA nos mercados e em finanças seja efetivo, especialmente no que se refere a proteção de informações sensíveis e financeiras dos participantes, e a proteção dos participantes dos riscos que o uso das informações providas por algoritmos podem trazer. O Fintech Times \footnote{\url{https://thefintechtimes.com/the-ethics-of-ai-ai-in-the-financial-services-sector-grand-opportunities-and-great-challenges/}} aponta três temas sensíveis que merecem atenção ao introduzir recursos de IA no setor financeiro e nos mercados em geral:

\begin{itemize}

\item{\textit{Ausência de Viés}: Antever que podem ocorrer falhas no projeto de algoritmos, e consequências indesejadas. Por exemplo, um algoritmo falho poderia adquirir um comportamento predatório ao considerar a necessidade da instituição financeira de otimizar sua lucratividade nas operações. Um algoritmo falho poderia se tornar predatório errando na estimativa da capacidade de um cliente de se endividar e de assumir riscos, e então oferecer ativos muito arriscados para este cliente que não teria condições de correr tais riscos nem de assumir as dívidas associadas a eles;}

\item{\textit{Responsabilização e Regulação}: É importante: a) regulamentar quais fontes de informações podem ser usadas e o uso correto destas informações, e b) definir antecipadamente de quem seria a responsabilidade se houverem consequências indesejadas de possíveis erros gerados por algoritmos, e como lidar com estas consequências (ex: informações incorretas ou inexatas podem induzir a erros na tomada de decisão e/ou em transações);}

\item{\textit{Transparência}: O que levou o algoritmo a tomar uma decisão, ou executar uma ação, deveria ser conhecido e rastreável.}

\end{itemize}

Também é importante chamar a atenção para o fato de que \textit{ algoritmos podem ser usados como armas}. Portanto, o uso de algoritmos para propósitos antiéticos, tais como roubo de informações sensíveis de clientes, deve ser evitado e responsabilizado.

\subsection{Tendências}

Ao consolidar tarefas e analisar dados de forma mais rápida e exata que os humanos, espera-se que o uso intensivo de IA economize mais de USD 1 trilhão para os bancos e instituições financeiras nos EUA até 2030\footnote{\url{https://www.processmaker.com/blog/why-ai-is-the-future-of-finance/}}. McKinsey Co.\footnote{\url{ https://www.mckinsey.com/industries/financial-services/our-insights/ai-bank-of-the-future-can-banks-meet-the-ai-challenge}} estima que o sistema financeiro deverá ser transformado pelas mudanças tecnológicas, e as instituições financeiras precisarão aumentar seus investimentos em tecnologia da informação e IA para atingir altos níveis de digitalização, com qualidade. Será uma questão de sobrevivência, pois McKinsey Co. também estima que mais de $78\%$ dos clientes jovens não iriam na agência física de uma instituição financeira se tivessem uma alternativa.  

%\section{Transportes} %nao descomentar aqui; o \section vem dentro do arquivo
\section{Mobilidade Urbana}
\label{sec:transportes}

A agenda em torno de cidades inteligentes tem como um dos focos a mobilidade urbana inteligente (uso racional dos diversos meios de transporte, integrando-os e adaptando-os à demanda). Existem diversas possibilidades em relação ao uso de IA em geral -- e aprendizado de máquina em particular -- em tal agenda.
O restante desta seção joga luz em alguns aspectos a respeito de como a IA vem contribuindo, e como seu papel se torna cada vez mais decisivo. Desta forma, um verdadeiro trânsito inteligente resultará de indivíduos, semáforos e veículos conectados e trabalhando em conjunto. Nesta visão, semáforos inteligentes são alimentados com informação a respeito do estado da rede de tráfego, sobre os sobre semáforos vizinhos, eventos imprevistos, e outras informações.

Uma explicação mais detalhada sobre sistemas de transporte e simulação de tráfego pode ser encontrada em \cite{Bazzan&Kluegl2013book,bazzan2021ea}. A seguir, serão abordados dois problemas centrais, os quais motivam diversas aplicações de IA. O primeiro se dá pelo lado da demanda (como se deslocar de A até B de maneira eficiente), enquanto que o segundo se refere ao lado da oferta (controle e gerenciamento de tráfego). Devido à limitação de espaço, nos concentramos no tráfego veicular urbano.

\subsection{Aplicações}

Em relação à oferta, quando se fala em mobilidade inteligente, as pessoas em geral pensam em semáforos inteligentes. 
Algoritmos e técnicas de controle semafórico existem há várias décadas e derivam principalmente de técnicas de pesquisa operacional e da área de controle. Mais recentemente, técnicas de IA e aprendizado de máquina têm sido empregadas, em especial aqueles que se baseiam em \rl\ (Seção~\ref{sec:rl}).
Neste caso, os semáforos devem aprender uma política que mapeia os estados (normalmente as filas nas interseções) para ações. Devido ao número de trabalhos que empregam \rl\ no controle semafórico, e às diversas modelagens e técnicas empregadas, sugere-se consultar os \textit{surveys} \cite{Bazzan2009,Wei+2019colight,Yau+2017,Noaeen+2022}.

Já pelo lado da demanda, entender como um motorista se comporta é fundamental em um sistema de recomendação de rotas e disseminação de informação aos motoristas. Não são muitos os trabalhos que consideram IA neste contexto. Redes neurais são utilizadas em 
\cite{Dia&Panwai2014} e em \cite{Barthelemy&Carletti2017} para prever e guiar, respectivamente, a escolha de rota dos motoristas. 
No caso da pesquisa mais recente, o foco é em: disseminação de informação, comunicação veicular, como aprender a escolher rotas, efeito de mudanças de comportamento da parte dos motoristas na presença de informação, e como disseminar informação de modo a garantir um determinado nível de desempenho do sistema.
Para atingir tais objetivos,  diversos métodos foram propostos no nosso grupo de pesquisa; para uma visão geral, ver \cite{Bazzan2022arxiv}.
Alguns destes trabalhos foram pioneiros ao abordar a disseminação de informação via dispositivos móveis quando o \textit{smartphone} não existia como o conhecemos hoje \cite{Kluegl&Bazzan2004jasss}. Outros métodos envolvem  comunicação interveicular \cite{Santos&Bazzan2021}, escolha de rota via \rl\ \cite{Bazzan&Grunitzki2016}, e efeito de recomendação de rotas \cite{Ramos+2018trc}.

Por fim, vale lembrar que também é possível utilizar IA em cenários que combinam controle semafórico com gerenciamento da demanda. De fato, esta integração, tão óbvia quanto importante, tem recebido pouca atenção na literatura. No trabalho de  \cite{Wiering2000} foi um dos pioneiros a tratar motoristas e semáforos aprendendo simultaneamente.
Em \cite{Lemos+2018} foi proposta uma abordagem baseada em jogos repetidos (para a classe motorista) e jogos estocásticos (para os semáforos). Por se tratar de naturezas  diversas de aprendizado, o artigo também discute os desafios encontrados em termos de \rl.

\subsection{Riscos}
De modo geral, os riscos de emprego de IA em aplicações na área de mobilidade urbana são similares a outras já discutidas neste capítulo.
Entretanto, entre algumas características específicas, destacam-se as seguintes. Em primeiro lugar, a área de controle semafórico tem a  segurança como absoluta prioridade. Desta forma, qualquer método de controle, seja ou não baseado em IA, deve fornecer garantias de que a sinalização não resultará em situações que violem os preceitos fundamentais.
Em segundo lugar, no que tange questões de comunicação interveicular, é obviamente fundamental garantir não apenas a privacidade dos envolvidos, mas também a segurança geral do sistema (por exemplo contra ataques maliciosos).

\subsection{Tendências}

Esta seção focou apenas nas questões anteriormente mencionadas -- a maioria relacionada a tráfego veicular urbano --. Entretanto, além das questões relacionadas a comunicação interveicular, existem pelo menos três áreas nas quais espera-se avanços significativos pelo uso da IA. A primeira está ligada a otimização do uso da rede de recarga de veículos elétricos. A segunda está, obviamente, relacionada com veículos autônomos e, principalmente, a como acomodar frotas mistas (autônomos e convencionais interagindo no mesmo ambiente). Por fim -- e em um horizonte mais concreto de tempo -- as aplicações de \textit{mobility as a service} já estão maduras e prontas para serem empregadas em conjuntos de políticas públicas visando dar acesso mais eficiente à populações cada vez mais heterogêneas em suas necessidades de mobilidade.  

%\section{IA Neuro-simbólica} %nao descomentar aqui; o \section vem dentro do arquivo
% Movido p/ conclusao
%\input{sections/neurosimb}
\part*{Parte III: Conclusões e Perspectivas}
%\section{Considerações finais}
\label{sec:conclusao}

%\subsection{Visão geral}
\section{Visão geral}
Este capítulo apresentou uma introdução aos fundamentos de IA e discutiu aplicações, riscos e tendências em suas múltiplas áreas. 
Destacamos que tal apresentação não é exaustiva. Há muitos outros conceitos envolvendo IA e muitas outras áreas impactadas que apenas tangenciamos.
Este capítulo pode ser visto como um ponto de partida, onde o leitor interessado poderá usar as referências apresentadas para se aprofundar nos tópicos de interesse.

Ao longo do capítulo, é possível ver que a aplicação da IA apresenta riscos em comum nas diferentes áreas. Especificamente, sistemas baseados em aprendizado supervisionado, incluindo aprendizado profundo,
possuem questões críticas relacionadas à semântica, explicabilidade, transparência (como e porquê determinada saída foi produzida) e vieses (resultados discriminatórios contra determinados grupos de pessoas).
Uma interpretação dos modelos de aprendizado é importante do ponto de vista tecnológico (e de produto) para oferecer garantias sobre o comportamento de um sistema. Ademais, entender exatamente porque um determinado sistema apresenta tal comportamento é um requisito básico de qualquer produto tecnológico.  
Nesse sentido, modelos de aprendizado profundo, embora tenham apresentados resultados tecnológicos relevantes, não apresentam uma semântica rigorosa (isto é, não são modelos que tenham associados uma interpretação lógico-matemática). 

Porém, riscos e acidentes não são exclusividade da IA. 
Todas as novidades tecnológicas da história da humanidade vieram com seus riscos. Como exemplos: junto com a introdução do automóvel vieram os acidentes automobilísticos e com a eletricidade, vieram os riscos de incêndios causados por curto-circuitos e acidentes por descarga elétrica, entre outros riscos que acompanham tecnologias. 
Uma questão importante é que nessas tecnologias, um acidente ou evento indesejado ocorre quando ``algo vai mal'', por falha humana, de hardware, de software, entre outras. Por exemplo, um acidente com automóvel ocorre por falha humana ou em algum de seus componentes; um curto-circuito ocorre por sobrecarga na fiação elétrica.
Em contraste, sistemas de IA baseados em aprendizado profundo tem a peculiaridade de que um evento indesejado ocorre mesmo quando ``tudo vai bem''. 
Mesmo com toda a implementação correta, e sem falhas no hardware, um sistema como o ChatGPT pode produzir saídas incorretas ou prejudiciais, conforme consta no próprio \textit{disclaimer} em sua página inicial\footnote{``Prévia de Pesquisa Gratuita. O ChatGPT pode produzir informações imprecisas sobre pessoas, lugares ou fatos. Versão do ChatGPT de 3 de Agosto'', conforme acesso em 22/09/2023.}.

Um grande tópico de pesquisa envolve, portanto, a identificação e mitigação desses riscos associados à IA. 
Alguns avanços foram feitos no caminho da explicabilidade \cite{Ribeiro2016lime,Lundberg2017shap} e na mitigação de vieses e outros riscos de segurança \cite{Thomas2019seldonian}.
Um promissor caminho integrador entre a IA baseada em aprendizado (eficiente, mas pouco transparente e por vezes pouco previsível) e a IA simbólica (menos eficiente até o momento, mas transparente, explicável e previsível) é a abordagem neuro-simbólica, cujos estudos visam, entre outros objetivos, oferecer interpretações (ou explicações, se a posteriori) dos métodos e mecanismos de aprendizado atualmente utilizados em IA, conforme discussão a seguir.

%\subsection{Integração para lidar com os desafios: a IA neuro-simbólica}
\section{Integração para lidar com os desafios: A IA Neuro-simbólica}
\label{sec:neurosimb}
%%Begin Lamb escreveu
Historicamente, a IA iniciou sua trajetória buscando a integração entre diversas habilidades cognitivas, dentre elas, o raciocínio e o aprendizado. Ambas dimensões são vistas como centrais à ideia de inteligência de máquina, já nos trabalhos originais de Turing, von Neumann, McCulloch, Pitts, entre outros \cite{turing1950}. von Neumann, em seus trabalhos iniciais, já identificava a relação entre a lógica intuicionista \cite{von1956probabilistic,GarcezLG06} e as redes neurais propostas por \cite{McCulloch1943}\footnote{%von Neumann foi explícito ao identificar a relação entre a lógica intuicionista e os modelos de rede neural de \cite{McCulloch1943}. 
\cite[Seção 2]{von1956probabilistic} afirma que "It has been pointed out by A. M. Turing [5] in 1937 and by W. S. McCulloch and W. Pitts [2] in 1943 that effectively constructive logics, that is, intuitionistic logics, can be best studied in terms of automata. Thus logical propositions can be represented as
electrical networks or (idealized) nervous systems." As referências [5] e [2] na citação são, respectivamente, \cite{Turing1937computability} e \cite{McCulloch1943}. }.

A área de IA neuro-simbólica integra os dois principais paradigmas da IA: conexionismo (notadamente associado ao uso de redes neurais artificiais como seu modelo principal) e simbolismo (onde o processo de raciocínio em IA é representado através de lógicas, incluindo diversas modalidades como tempo, espaço, conhecimento e incerteza) \cite{GarcezLG07,AAAI07,3rdWave}. 
%Na história recente, a partir dos anos 1990, 
Tradicionalmente, estas áreas foram desenvolvidas por correntes diversas, por terem fundamentos computacionais e lógicos distintos \cite{Besold2022,GarcezLG2009}. 
A área recebeu certa atenção inicial nos anos 1990 e 2000 \cite{hinton}, quando pesquisadores passaram a desenvolver abordagens  neuro-simbólicas que aprendessem a realizar inferência lógica clássica, mesmo que para fragmentos de lógicas de predicados \cite{Audibert2022,Garcez_1999}. À época, foram desenvolvidos sistemas neuro-simbólicos que aprendiam a computar (fragmentos) de programas escritos em linguagens lógicas, como Prolog. Posteriormente, pesquisadores demonstraram que modelos conexionistas poderiam ser treinados para aprender regras de inferência lógica, notadamente sobre lógicas não-clássicas, permitindo a expressão de multi-modalidades \cite{NIPS03,NECO3050,AAAI07}, antecipando, de certa forma, a pesquisa atual em grandes modelos de linguagens que visa expressar multimodalidades na interação entre usuários humanos e esses sistemas \cite{Kiros2014}. 

As grandes contribuições que a área de IA neuro-simbólica pode oferecer são sumarizadas em artigos recentes, publicados na \emph{Communications of the ACM} \cite{Monroe2022, Hochreiter22}. Monroe ressalta a necessidade de desenvolvimento de uma semântica rigorosa para os modelos de IA, como defendido em \cite{3rdWave,LambIJCAI2020}, enquanto Hochreiter aponta que a forma de desenvolver uma IA ampla, que contemple múltiplas habilidades cognitivas, pode ser melhor atingida através da IA neuro-simbólica, sugerindo a abordagem de redes grafos neurais neuro-simbólicos. Hochreiter cita especificamente o trabalho \cite{LambIJCAI2020} como sendo promissor para esta linha de pesquisa em IA ampla. %, que vai além dos sistemas e tecnologias atuais, que somente são focados em um dimensão cognitiva, como e.g., reconhecimento de imagens, visão computacional, ou processamento de linguagem natural.  
É relevante ressaltar que esta necessidade de integração neuro-simbólica foi apontada como promissora em eventos recentes, como nas conferências AAAI e NeurIPS, bem como nos debates organizados pela  Montreal AI, denominadas de "AI Debates"\footnote{\url{https://www.quebecartificialintelligence.com/aidebate2/}} números 1, 2 e 3.  Nestes eventos, foi apontado que para construir sistemas que representem as duas formas de raciocínio -  na forma de "AI Fast and Slow" inspirado em \cite{Kahneman2017thinking} (ver Seção \ref{sec:fundamentos}) - visando a integração de múltiplas habilidades cognitivas em IA, possíveis abordagens promissoras seriam justamente as adotadas na IA neuro-simbólica. Estas abordagens oferecem a possibilidade do desenvolvimento de modelos com fundamentação rigorosa e transparente do ponto de vista lógico, que se integrados ao aprendizado profundo podem levar a tecnologias de inteligência artificial mais robustas, explicáveis e transparentes, oferecendo maior segurança a todos que fazem uso desta tecnologia de propósito geral \cite{3rdWave}.

Mais informações sobre a evolução da IA, em particular IA neuro-simbólica podem ser encontradas em \cite{Audibert2022,GarcezLG2009,3rdWave}.

%\subsection{Epílogo}
\section{Epílogo}
\label{sec:epilogo}
Os impactos sociais e éticos da IA levantam questões preponderantes no debate científico e na grande mídia. Atualmente, inúmeras organizações científicas\footnote{e.g. AAAI (Association for the Advancement of Artificial Intelligence), ACM (Association for Computing Machinery), IEEE (Institute of Electrical and Electronic Engineers), e Royal Society, entre outras.} e academias nacionais de ciência têm debatido o impacto da IA na ciência e na sociedade. 
Além das entidades científicas, organismos multilaterais, historicamente dedicados a temáticas econômicas e sociais\footnote{e.g. Fórum Econômico Mundial (WEF), Organização para Cooperação e Desenvolvimento Econômico (OCDE), e as Nações Unidas.}, constituíram grupos de trabalho sobre o impacto da IA. %Vários países criaram grupos para definir estratégias nacionais de IA, bem como as atualizaram, tendo em vista o impacto recente da área sobre a vida humana, a economia e o futuro do planeta. 
Muitas vezes este debate é alimentado por não-especialistas, o que demanda cuidados na interpretação do que é publicado. Mesmo com esta ressalva, é muito relevante perceber que nos últimos 10 a 15 anos, com o crescente impacto da IA, particularmente do aprendizado profundo, os cientistas identificaram uma série de limitações e preocupações com o uso da IA sem curadoria - i.e., sem a própria análise de uso por especialistas.
%Isso deixa claro que a IA não é apenas uma tecnologia do futuro, mas já se faz presente. 

%À medida que a IA continua a revolucionar setores, sociedades e a maneira como entendemos a própria inteligência, enfrentamos desafios cruciais de ética, privacidade e responsabilidade.
%Com uma compreensão humanista e uma abordagem orientada por valores, podemos aproveitar o potencial da IA para melhoria de qualidade de vida, abordando importantes desafios como crescimento populacional, meio ambiente e sustentabilidade. 

Por fim, deixamos uma demonstração da capacidade utilitária da IA e como ela pode potencializar a capacidade criativa dos seres humanos. Em resposta ao seguinte \emph{prompt}: ``escreva um pequeno texto de conclusão para este capítulo de livro ``A Nova Eletricidade'': Aplicações, Riscos e Tendências da IA Moderna'', a auspiciosa última frase do texto gerado pelo ChatGPT foi: ``À medida que fechamos este capítulo, é imperativo olhar para o horizonte da IA com olhos críticos e curiosos, prontos para navegar nas águas emocionantes, porém desafiadoras, deste novo mundo alimentado por algoritmos e dados. O futuro da IA está nas mãos daqueles que a guiam com sabedoria e visão.'' 

Cabe aos seres humanos, portanto, a contínua busca por sabedoria e visão para guiar a IA e todas as tecnologias presentes e futuras para o próprio bem da humanidade.

\section*{Agradecimentos}

O presente trabalho foi realizado com apoio da Coordenação de Aperfeiçoamento de Pessoal de Nível Superior - Brasil (CAPES) - Código de Financiamento 001 e CNPq – Conselho Nacional de Desenvolvimento Científico e Tecnológico.
Agradecimentos também a Cláudio Geyer por comentários no texto e ajuda na revisão.

\bibliographystyle{sbc}
\begin{small} %for 9pt size 
\bibliography{refs,AL,MZ,OURS}

\begin{thebibliography}{}

\bibitem[Agrawal et~al. 2019]{agrawal2019artificial}
Agrawal, A., Gans, J.~S., and Goldfarb, A. (2019).
\newblock Artificial intelligence: the ambiguous labor market impact of
  automating prediction.
\newblock {\em Journal of Economic Perspectives}, 33(2):31--50.

\bibitem[Al-Zaiti et~al. 2023]{al2023machine}
Al-Zaiti, S.~S., Martin-Gill, C., Z{\`e}gre-Hemsey, J.~K., Bouzid, Z.,
  Faramand, Z., Alrawashdeh, M.~O., Gregg, R.~E., Helman, S., Riek, N.~T.,
  Kraevsky-Phillips, K., et~al. (2023).
\newblock Machine learning for {ECG} diagnosis and risk stratification of
  occlusion myocardial infarction.
\newblock {\em Nature Medicine}, pages 1--10.

\bibitem[Aljameel et~al. 2022]{aljameel2022anomaly}
Aljameel, S.~S., Alomari, D.~M., Alismail, S., Khawaher, F., Alkhudhair, A.~A.,
  Aljubran, F., and Alzannan, R.~M. (2022).
\newblock An anomaly detection model for oil and gas pipelines using machine
  learning.
\newblock {\em Computation}, 10(8):138.

\bibitem[Aphirakmethawong et~al. 2022]{aphirakmethawong2022overview}
Aphirakmethawong, J., Yang, E., and Mehnen, J. (2022).
\newblock An overview of artificial intelligence in product design for smart
  manufacturing.
\newblock In {\em 2022 27th International Conference on Automation and
  Computing (ICAC)}, pages 1--6. IEEE.

\bibitem[Audibert et~al. 2022]{Audibert2022}
Audibert, R.~B., dos Santos, H.~L., Avelar, P. H.~C., Tavares, A.~R., and Lamb,
  L.~C. (2022).
\newblock On the evolution of {A.I.} and machine learning: Towards measuring
  and understanding impact, influence, and leadership at premier {A.I.}
  conferences.
\newblock {\em arXiv preprint arXiv:2205.13131}.

\bibitem[Barth{\'e}lemy and Carletti 2017]{Barthelemy&Carletti2017}
Barth{\'e}lemy, J. and Carletti, T. (2017).
\newblock A dynamic behavioural traffic assignment model with strategic agents.
\newblock {\em Transportation Research Part C: Emerging Technologies},
  85:23--46.

\bibitem[Baum et~al. 2021]{baum2021artificial}
Baum, Z.~J., Yu, X., Ayala, P.~Y., Zhao, Y., Watkins, S.~P., and Zhou, Q.
  (2021).
\newblock Artificial intelligence in chemistry: current trends and future
  directions.
\newblock {\em Journal of Chemical Information and Modeling}, 61(7):3197--3212.

\bibitem[Bazzan 2021]{bazzan2021ea}
Bazzan, A.~L. (2021).
\newblock Contribui\c c\~oes de aprendizado por refor\c co em escolha de rota e
  controle semaf\'orico.
\newblock {\em Estudos Avan\c cados}, 35(101):95--110.

\bibitem[Bazzan 2009]{Bazzan2009}
Bazzan, A. L.~C. (2009).
\newblock Opportunities for multiagent systems and multiagent reinforcement
  learning in traffic control.
\newblock {\em Autonomous Agents and Multiagent Systems}, 18(3):342--375.

\bibitem[Bazzan 2022]{Bazzan2022arxiv}
Bazzan, A. L.~C. (2022).
\newblock Improving urban mobility: using artificial intelligence and new
  technologies to connect supply and demand.
\newblock \url{https://arxiv.org/abs/2204.03570}.

\bibitem[Bazzan and Grunitzki 2016]{Bazzan&Grunitzki2016}
Bazzan, A. L.~C. and Grunitzki, R. (2016).
\newblock A multiagent reinforcement learning approach to en-route trip
  building.
\newblock In {\em 2016 International Joint Conference on Neural Networks
  (IJCNN)}, pages 5288--5295.

\bibitem[Bazzan and Kl\"ugl 2013]{Bazzan&Kluegl2013book}
Bazzan, A. L.~C. and Kl\"ugl, F. (2013).
\newblock {\em Introduction to Intelligent Systems in Traffic and
  Transportation}, volume~7 of {\em Synthesis Lectures on Artificial
  Intelligence and Machine Learning}.
\newblock Morgan and Claypool.

\bibitem[Bender et~al. 2021]{bender2021dangers}
Bender, E.~M., Gebru, T., McMillan-Major, A., and Shmitchell, S. (2021).
\newblock On the dangers of stochastic parrots: Can language models be too big?
\newblock In {\em Proceedings of the 2021 ACM conference on fairness,
  accountability, and transparency}, pages 610--623.

\bibitem[Benos et~al. 2021]{benos2021machine}
Benos, L., Tagarakis, A.~C., Dolias, G., Berruto, R., Kateris, D., and Bochtis,
  D. (2021).
\newblock Machine learning in agriculture: A comprehensive updated review.
\newblock {\em Sensors}, 21(11):3758.

\bibitem[Berner et~al. 2019]{Berner2019dota}
Berner, C., Brockman, G., Chan, B., Cheung, V., D{\k{e}}biak, P., Dennison, C.,
  Farhi, D., Fischer, Q., Hashme, S., Hesse, C., et~al. (2019).
\newblock Dota 2 with large scale deep reinforcement learning.
\newblock {\em arXiv preprint arXiv:1912.06680}.

\bibitem[Besold et~al. 2022]{Besold2022}
Besold, T.~R., d'Avila Garcez, A.~S., Bader, S., Bowman, H., Domingos, P.~M.,
  Hitzler, P., K{\"{u}}hnberger, K., Lamb, L.~C., Lima, P. M.~V., de~Penning,
  L., Pinkas, G., Poon, H., and Zaverucha, G. (2022).
\newblock Neural-symbolic learning and reasoning: {A} survey and
  interpretation.
\newblock In Hitzler, P. and Sarker, M.~K., editors, {\em Neuro-Symbolic
  Artificial Intelligence: The State of the Art}, volume 342 of {\em Frontiers
  in Artificial Intelligence and Applications}, pages 1--51. {IOS} Press.

\bibitem[Birlutiu et~al. 2017]{birlutiu2017defect}
Birlutiu, A., Burlacu, A., Kadar, M., and Onita, D. (2017).
\newblock Defect detection in porcelain industry based on deep learning
  techniques.
\newblock In {\em 2017 19th International Symposium on Symbolic and Numeric
  Algorithms for Scientific Computing (SYNASC)}, pages 263--270. IEEE.

\bibitem[Brachman and Levesque 2004]{brachman2004knowledge}
Brachman, R.~J. and Levesque, H.~J. (2004).
\newblock {\em Knowledge Representation and Reasoning}.
\newblock Elsevier.

\bibitem[Broda et~al. 2004]{Broda2004}
Broda, K., Gabbay, D., Lamb, L., and Russo, A. (2004).
\newblock {\em Compiled Labelled Deductive Systems: A Uniform Presentation of
  Non-Classical Logics}.
\newblock Institute of Physics/Research Studies Press, Hertfordshire.

\bibitem[Brown et~al. 2020]{brown2020language}
Brown, T.~B., Mann, B., Ryder, N., Subbiah, M., Kaplan, J., Dhariwal, P.,
  Neelakantan, A., Shyam, P., Sastry, G., Askell, A., Agarwal, S.,
  Herbert-Voss, A., Krueger, G., Henighan, T., Child, R., Ramesh, A., Ziegler,
  D.~M., Wu, J., Winter, C., Hesse, C., Chen, M., Sigler, E., Litwin, M., Gray,
  S., Chess, B., Clark, J., Berner, C., McCandlish, S., Radford, A., Sutskever,
  I., and Amodei, D. (2020).
\newblock {Language Models are Few-Shot Learners}.
\newblock {\em arXiv preprint arXiv:2005.14165}.

\bibitem[Brownstein et~al. 2023]{brownstein2023advances}
Brownstein, J.~S., Rader, B., Astley, C.~M., and Tian, H. (2023).
\newblock Advances in artificial intelligence for infectious-disease
  surveillance.
\newblock {\em New England Journal of Medicine}, 388(17):1597--1607.

\bibitem[Buolamwini and Gebru 2018]{buolamwini2018gender}
Buolamwini, J. and Gebru, T. (2018).
\newblock Gender shades: Intersectional accuracy disparities in commercial
  gender classification.
\newblock In {\em Conference on fairness, accountability and transparency},
  pages 77--91. PMLR.

\bibitem[Colombelli et~al. 2022]{colombelli2022hybrid}
Colombelli, F., Kowalski, T.~W., and Recamonde-Mendoza, M. (2022).
\newblock A hybrid ensemble feature selection design for candidate biomarkers
  discovery from transcriptome profiles.
\newblock {\em Knowledge-Based Systems}, 254:109655.

\bibitem[Dalzochio et~al. 2020]{dalzochio2020machine}
Dalzochio, J., Kunst, R., Pignaton, E., Binotto, A., Sanyal, S., Favilla, J.,
  and Barbosa, J. (2020).
\newblock Machine learning and reasoning for predictive maintenance in industry
  4.0: Current status and challenges.
\newblock {\em Computers in Industry}, 123:103298.

\bibitem[{d'Avila Garcez} and Lamb 2006]{NECO3050}
{d'Avila Garcez}, A. and Lamb, L. (2006).
\newblock A connectionist computational model for epistemic and temporal
  reasoning.
\newblock {\em Neur. Computation}, 18(7):1711--1738.

\bibitem[{d'Avila Garcez} et~al. 2006]{GarcezLG06}
{d'Avila Garcez}, A., Lamb, L., and Gabbay, D. (2006).
\newblock Connectionist computations of intuitionistic reasoning.
\newblock {\em Theor. Comput. Sci.}, 358(1):34--55.

\bibitem[d'Avila Garcez and Lamb 2023]{3rdWave}
d'Avila Garcez, A. and Lamb, L.~C. (2023).
\newblock Neurosymbolic {AI}: {T}he 3rd {Wa}ve.
\newblock {\em Artificial Intelligence Review}.

\bibitem[d'Avila Garcez et~al. 2007]{GarcezLG07}
d'Avila Garcez, A., Lamb, L.~C., and Gabbay, D.~M. (2007).
\newblock Connectionist modal logic: Representing modalities in neural
  networks.
\newblock {\em Theor. Comput. Sci.}, 371(1-2):34--53.

\bibitem[d'Avila Garcez and Zaverucha 1999]{Garcez_1999}
d'Avila Garcez, A. and Zaverucha, G. (1999).
\newblock The connectionist inductive learning and logic programming system.
\newblock {\em Applied Intelligence}, 11(1):59–77.

\bibitem[d'Avila Garcez and Lamb 2003]{NIPS03}
d'Avila Garcez, A.~S. and Lamb, L.~C. (2003).
\newblock Reasoning about {T}ime and {K}nowledge in {N}eural-symbolic
  {L}earning {S}ystems.
\newblock In Thrun, S., Saul, L.~K., and Sch{\"{o}}lkopf, B., editors, {\em
  Advances in Neural Information Processing Systems 16 [Neural Information
  Processing Systems, {NIPS} 2003, December 8-13, 2003, Vancouver and Whistler,
  British Columbia, Canada]}, pages 921--928. {MIT} Press.

\bibitem[d'Avila Garcez et~al. 2009]{GarcezLG2009}
d'Avila Garcez, A.~S., Lamb, L.~C., and Gabbay, D.~M. (2009).
\newblock {\em Neural-Symbolic Cognitive Reasoning}.
\newblock Cognitive Technologies. Springer.

\bibitem[Devlin et~al. 2019]{devlin2018bert}
Devlin, J., Chang, M.-W., Lee, K., and Toutanova, K. (2019).
\newblock {BERT: Pre-training of Deep Bidirectional Transformers for Language
  Understanding}.
\newblock {\em arXiv preprint arXiv:1810.04805}.

\bibitem[Dia and Panwai 2014]{Dia&Panwai2014}
Dia, H. and Panwai, S. (2014).
\newblock {\em Intelligent Transport Systems: Neural Agent (Neugent) Models of
  Driver Behaviour}.
\newblock LAP Lambert Academic Publishing.

\bibitem[dos Anjos et~al. 2023]{Anjos-ACM-092023}
dos Anjos, J. C.~S., Matteussi, K.~J., Orlandi, F.~C., Barbosa, J. L.~V.,
  Silva, J.~S., Bittencourt, L.~F., and Geyer, C. F.~R. (2023).
\newblock {A Survey on Collaborative Learning for Intelligent Autonomous
  Systems}.
\newblock {\em ACM Comput. Surv.}, 1(1):1–36.

\bibitem[Dosovitskiy et~al. 2020]{dosovitskiy2020image}
Dosovitskiy, A., Beyer, L., Kolesnikov, A., Weissenborn, D., Zhai, X.,
  Unterthiner, T., Dehghani, M., Minderer, M., Heigold, G., Gelly, S., et~al.
  (2020).
\newblock An image is worth 16x16 words: Transformers for image recognition at
  scale.
\newblock {\em arXiv preprint arXiv:2010.11929}.

\bibitem[Eykholt et~al. 2018]{eykholt2018robust}
Eykholt, K., Evtimov, I., Fernandes, E., Li, B., Rahmati, A., Xiao, C.,
  Prakash, A., Kohno, T., and Song, D. (2018).
\newblock Robust physical-world attacks on deep learning visual classification.
\newblock In {\em Proceedings of the IEEE conference on computer vision and
  pattern recognition}, pages 1625--1634.

\bibitem[Fagin et~al. 1995]{Fagin1995}
Fagin, R., Halpern, J.~Y., Moses, Y., and Vardi, M.~Y. (1995).
\newblock {\em Reasoning About Knowledge}.
\newblock {MIT} Press.

\bibitem[Fournier-Viger et~al. 2021]{fournier2021machine}
Fournier-Viger, P., Nawaz, M.~S., Song, W., and Gan, W. (2021).
\newblock Machine learning for intelligent industrial design.
\newblock In {\em Joint European Conference on Machine Learning and Knowledge
  Discovery in Databases}, pages 158--172. Springer.

\bibitem[Fragapane et~al. 2022]{fragapane2022increasing}
Fragapane, G., Ivanov, D., Peron, M., Sgarbossa, F., and Strandhagen, J.~O.
  (2022).
\newblock Increasing flexibility and productivity in industry 4.0 production
  networks with autonomous mobile robots and smart intralogistics.
\newblock {\em Annals of Operations Research}, 308(1-2):125--143.

\bibitem[Franklin et~al. 2020]{franklin2020collaborative}
Franklin, C.~S., Dominguez, E.~G., Fryman, J.~D., and Lewandowski, M.~L.
  (2020).
\newblock Collaborative robotics: New era of human--robot cooperation in the
  workplace.
\newblock {\em Journal of Safety Research}, 74:153--160.

\bibitem[Geffner 2018]{geffner2018model}
Geffner, H. (2018).
\newblock Model-free, model-based, and general intelligence.
\newblock In {\em Proceedings of the Twenty-Seventh International Joint
  Conference on Artificial Intelligence, {IJCAI}2018}.

\bibitem[Giri et~al. 2019]{giri2019detailed}
Giri, C., Jain, S., Zeng, X., and Bruniaux, P. (2019).
\newblock A detailed review of artificial intelligence applied in the fashion
  and apparel industry.
\newblock {\em IEEE Access}, 7:95376--95396.

\bibitem[Goodfellow et~al. 2016]{Goodfellow2016deeplearning}
Goodfellow, I., Bengio, Y., and Courville, A. (2016).
\newblock {\em Deep {L}earning}.
\newblock {MIT} Press.

\bibitem[Grisoni et~al. 2021]{grisoni2021combining}
Grisoni, F., Huisman, B.~J., Button, A.~L., Moret, M., Atz, K., Merk, D., and
  Schneider, G. (2021).
\newblock Combining generative artificial intelligence and on-chip synthesis
  for de novo drug design.
\newblock {\em Science Advances}, 7(24):eabg3338.

\bibitem[Gubbi et~al. 2022]{gubbi2022survey}
Gubbi, K.~I., Beheshti-Shirazi, S.~A., Sheaves, T., Salehi, S., PD, S.~M.,
  Rafatirad, S., Sasan, A., and Homayoun, H. (2022).
\newblock Survey of machine learning for electronic design automation.
\newblock In {\em Proceedings of the Great Lakes Symposium on VLSI 2022}, pages
  513--518.

\bibitem[Hamolia and Melnyk 2021]{hamolia2021survey}
Hamolia, V. and Melnyk, V. (2021).
\newblock A survey of machine learning methods and applications in electronic
  design automation.
\newblock In {\em 2021 11th International Conference on Advanced Computer
  Information Technologies (ACIT)}, pages 757--760. IEEE.

\bibitem[Hart et~al. 1968]{Hart1968}
Hart, P.~E., Nilsson, N.~J., and Raphael, B. (1968).
\newblock A formal basis for the heuristic determination of minimum cost paths.
\newblock {\em IEEE Transactions on Systems Science and Cybernetics},
  4(2):100--107.

\bibitem[Hasan et~al. 2021]{hasan2021generalizable}
Hasan, I., Liao, S., Li, J., Akram, S.~U., and Shao, L. (2021).
\newblock Generalizable pedestrian detection: The elephant in the room.
\newblock In {\em Proceedings of the IEEE/CVF Conference on Computer Vision and
  Pattern Recognition}, pages 11328--11337.

\bibitem[Helmert and Domshlak 2009]{Helmert2009}
Helmert, M. and Domshlak, C. (2009).
\newblock Landmarks, critical paths and abstractions: What's the difference
  anyway?
\newblock In {\em {International Conference on Automated Planning and
  Scheduling}}, pages 162--169.

\bibitem[Hinton 1990]{hinton}
Hinton, G. (1990).
\newblock Connectionist symbol processing - preface.
\newblock {\em Artif. Intell.}, 46(1-2):1--4.

\bibitem[Hinton et~al. 2006]{Hinton_2006}
Hinton, G.~E., Osindero, S., and Teh, Y.-W. (2006).
\newblock A fast learning algorithm for deep belief nets.
\newblock {\em Neural Computation}, 18(7):1527–1554.

\bibitem[Hochreiter 2022]{Hochreiter22}
Hochreiter, S. (2022).
\newblock Toward a broad {AI}.
\newblock {\em Communications of the {ACM}}, 65(4):56--57.

\bibitem[Hochreiter and Schmidhuber 1997]{hochreiter1997long}
Hochreiter, S. and Schmidhuber, J. (1997).
\newblock Long short-term memory.
\newblock {\em Neural Computation}, 9(8):1735--1780.

\bibitem[Hoffmann 2011]{hoffmann2011everything}
Hoffmann, J. (2011).
\newblock Everything you always wanted to know about planning: (but were afraid
  to ask).
\newblock In {\em Advances in Artificial Intelligence}, pages 1--13.

\bibitem[Huang et~al. 2019]{huang2019prediction}
Huang, P., Lin, C.~T., Li, Y., Tammemagi, M.~C., Brock, M.~V., Atkar-Khattra,
  S., Xu, Y., Hu, P., Mayo, J.~R., Schmidt, H., et~al. (2019).
\newblock Prediction of lung cancer risk at follow-up screening with low-dose
  {CT}: a training and validation study of a deep learning method.
\newblock {\em The Lancet Digital Health}, 1(7):e353--e362.

\bibitem[Ignat et~al. 2023]{ignat2023phd}
Ignat, O., Jin, Z., Abzaliev, A., Biester, L., Castro, S., Deng, N., Gao, X.,
  Gunal, A., He, J., Kazemi, A., Khalifa, M., Koh, N., Lee, A., Liu, S., Min,
  D.~J., Mori, S., Nwatu, J., Perez-Rosas, V., Shen, S., Wang, Z., Wu, W., and
  Mihalcea, R. (2023).
\newblock A {P}h{D} student's perspective on research in {NLP} in the era of
  very large language models.

\bibitem[Jiang et~al. 2018]{jiang2018mapping}
Jiang, D., Hao, M., Ding, F., Fu, J., and Li, M. (2018).
\newblock Mapping the transmission risk of zika virus using machine learning
  models.
\newblock {\em Acta Tropica}, 185:391--399.

\bibitem[Jumper et~al. 2021]{jumper2021highly}
Jumper, J., Evans, R., Pritzel, A., Green, T., Figurnov, M., Ronneberger, O.,
  Tunyasuvunakool, K., Bates, R., {\v{Z}}{\'\i}dek, A., Potapenko, A., et~al.
  (2021).
\newblock Highly accurate protein structure prediction with {AlphaFold}.
\newblock {\em Nature}, 596(7873):583--589.

\bibitem[Kahneman 2011]{Kahneman2017thinking}
Kahneman, D. (2011).
\newblock {\em Thinking, {F}ast and {S}low}.
\newblock Farrar, Straus and Giroux.

\bibitem[Kielarova and Pradujphongphet 2023]{kielarova2023genetic}
Kielarova, S.~W. and Pradujphongphet, P. (2023).
\newblock Genetic algorithm for product design optimization: An industrial case
  study of halo setting for jewelry design.
\newblock In {\em International Conference on Swarm Intelligence}, pages
  219--228. Springer.

\bibitem[Kiros et~al. 2014]{Kiros2014}
Kiros, R., Salakhutdinov, R., and Zemel, R. (2014).
\newblock Multimodal neural language models.
\newblock In {\em Proceedings of the 31st International Conference on
  International Conference on Machine Learning - Volume 32}, ICML'14, page
  II–595–II–603. JMLR.org.

\bibitem[Kl\"ugl and Bazzan 2004]{Kluegl&Bazzan2004jasss}
Kl\"ugl, F. and Bazzan, A. L.~C. (2004).
\newblock Route decision behaviour in a commuting scenario.
\newblock {\em Journal of Artificial Societies and Social Simulation}, 7(1).

\bibitem[Kowalski 1979]{Kowalski79}
Kowalski, R.~A. (1979).
\newblock {\em Logic for problem solving}.
\newblock North-Holland.

\bibitem[Lamb et~al. 2007]{AAAI07}
Lamb, L., Borges, R., and {d'Avila Garcez}, A. (2007).
\newblock A connectionist cognitive model for temporal synchronisation and
  learning.
\newblock In {\em Proceedings of the 22nd National Conference on Artificial
  Intelligence - Volume 1}, AAAI'07, page 827–832.

\bibitem[Lamb et~al. 2020]{LambIJCAI2020}
Lamb, L.~C., d'Avila Garcez, A.~S., Gori, M., Prates, M. O.~R., Avelar, P.
  H.~C., and Vardi, M.~Y. (2020).
\newblock Graph neural networks meet neural-symbolic computing: {A} survey and
  perspective.
\newblock In Bessiere, C., editor, {\em Proceedings of the Twenty-Ninth
  International Joint Conference on Artificial Intelligence, {IJCAI} 2020},
  pages 4877--4884. ijcai.org.

\bibitem[LeCun et~al. 1989]{LeCun1989cnn}
LeCun, Y., Boser, B., Denker, J.~S., Henderson, D., Howard, R.~E., Hubbard, W.,
  and Jackel, L.~D. (1989).
\newblock {Backpropagation Applied to Handwritten Zip Code Recognition}.
\newblock {\em Neural Computation}, 1(4):541--551.

\bibitem[Lemos et~al. 2018]{Lemos+2018}
Lemos, L.~L., Bazzan, A. L.~C., and Pasin, M. (2018).
\newblock Co-adaptive reinforcement learning in microscopic traffic systems.
\newblock In {\em 2018 {IEEE} Congress on Evolutionary Computation, {CEC} 2018,
  Rio de Janeiro, Brazil, July 8-13, 2018}, pages 1--8.

\bibitem[Li et~al. 2017]{li2017applications}
Li, B.-h., Hou, B.-c., Yu, W.-t., Lu, X.-b., and Yang, C.-w. (2017).
\newblock Applications of artificial intelligence in intelligent manufacturing:
  a review.
\newblock {\em Frontiers of Information Technology \& Electronic Engineering},
  18:86--96.

\bibitem[Liang et~al. 2020]{liang2020implementation}
Liang, Y., Lee, S.-H., and Workman, J.~E. (2020).
\newblock Implementation of artificial intelligence in fashion: Are consumers
  ready?
\newblock {\em Clothing and Textiles Research Journal}, 38(1):3--18.

\bibitem[Liu et~al. 2019]{liu2019comparison}
Liu, X., Faes, L., Kale, A.~U., Wagner, S.~K., Fu, D.~J., Bruynseels, A.,
  Mahendiran, T., Moraes, G., Shamdas, M., Kern, C., et~al. (2019).
\newblock A comparison of deep learning performance against health-care
  professionals in detecting diseases from medical imaging: a systematic review
  and meta-analysis.
\newblock {\em The Lancet Digital Health}, 1(6):e271--e297.

\bibitem[Lundberg and Lee 2017]{Lundberg2017shap}
Lundberg, S.~M. and Lee, S.-I. (2017).
\newblock A unified approach to interpreting model predictions.
\newblock In Guyon, I., Luxburg, U.~V., Bengio, S., Wallach, H., Fergus, R.,
  Vishwanathan, S., and Garnett, R., editors, {\em Advances in Neural
  Information Processing Systems 30}, pages 4765--4774. Curran Associates, Inc.

\bibitem[Lynch 2017]{Lynch2017andrewng}
Lynch, S. (2017).
\newblock {{A}ndrew {N}g: Why {AI} Is the New Electricity}.
\newblock
  \url{https://www.gsb.stanford.edu/insights/andrew-ng-why-ai-new-electricity}.
\newblock Acesso em 15/09/2023.

\bibitem[Mahendran and PM 2022]{mahendran2022deep}
Mahendran, N. and PM, D. R.~V. (2022).
\newblock A deep learning framework with an embedded-based feature selection
  approach for the early detection of the {Alzheimer}'s disease.
\newblock {\em Computers in Biology and Medicine}, 141:105056.

\bibitem[Marczyk et~al. 2023]{marczyk2023classification}
Marczyk, V.~R., Recamonde-Mendoza, M., Maia, A.~L., and Goemann, I.~M. (2023).
\newblock Classification of thyroid tumors based on {DNA} methylation patterns.
\newblock {\em Thyroid}, 33(9):1090--1099.

\bibitem[Mart{\'\i} et~al. 2015]{marti2015anomaly}
Mart{\'\i}, L., Sanchez-Pi, N., Molina, J.~M., and Garcia, A. C.~B. (2015).
\newblock Anomaly detection based on sensor data in petroleum industry
  applications.
\newblock {\em Sensors}, 15(2):2774--2797.

\bibitem[Martin et~al. 2019]{martin2019clinical}
Martin, A.~R., Kanai, M., Kamatani, Y., Okada, Y., Neale, B.~M., and Daly,
  M.~J. (2019).
\newblock Clinical use of current polygenic risk scores may exacerbate health
  disparities.
\newblock {\em Nature Genetics}, 51(4):584--591.

\bibitem[McCulloch and Pitts 1943]{McCulloch1943}
McCulloch, W.~S. and Pitts, W. (1943).
\newblock A logical calculus of the ideas immanent in nervous activity.
\newblock {\em The Bulletin of Mathematical Biophysics}, 5(4):115--133.

\bibitem[Mikolov et~al. 2013]{Mikolov2013word2vec}
Mikolov, T., Chen, K., Corrado, G., and Dean, J. (2013).
\newblock Efficient estimation of word representations in vector space.
\newblock {\em arXiv preprint arXiv:1301.3781}.

\bibitem[Mitchell 1997]{Mitchell1997}
Mitchell, T.~M. (1997).
\newblock {\em Machine Learning}.
\newblock McGraw-Hill, Inc., New York, NY, USA, 1 edition.

\bibitem[Mnih et~al. 2015]{Mnih+2015}
Mnih, V., Kavukcuoglu, K., Silver, D., Rusu, A.~A., Veness, J., Bellemare,
  M.~G., Graves, A., Riedmiller, M., Fidjeland, A.~K., Ostrovski, G., et~al.
  (2015).
\newblock Human-level control through deep reinforcement learning.
\newblock {\em Nature}, 518(7540):529--533.

\bibitem[Monroe 2022]{Monroe2022}
Monroe, D. (2022).
\newblock Neurosymbolic {AI}.
\newblock {\em Communications of the {ACM}}, 65(10):11–13.

\bibitem[Narodytska and Kasiviswanathan 2017]{narodytska2017simple}
Narodytska, N. and Kasiviswanathan, S.~P. (2017).
\newblock Simple black-box adversarial attacks on deep neural networks.
\newblock In {\em CVPR Workshops}, volume~2, page~2.

\bibitem[Noaeen et~al. 2022]{Noaeen+2022}
Noaeen, M., Naik, A., Goodman, L., Crebo, J., Abrar, T., Abad, Z. S.~H.,
  Bazzan, A.~L., and Far, B. (2022).
\newblock Reinforcement learning in urban network traffic signal control: A
  systematic literature review.
\newblock {\em Expert Systems with Applications}, page 116830.

\bibitem[Obermeyer et~al. 2019]{obermeyer2019dissecting}
Obermeyer, Z., Powers, B., Vogeli, C., and Mullainathan, S. (2019).
\newblock Dissecting racial bias in an algorithm used to manage the health of
  populations.
\newblock {\em Science}, 366(6464):447--453.

\bibitem[Ouyang et~al. 2022]{Ouyang2022instructgpt}
Ouyang, L., Wu, J., Jiang, X., Almeida, D., Wainwright, C.~L., Mishkin, P.,
  Zhang, C., Agarwal, S., Slama, K., Ray, A., Schulman, J., Hilton, J., Kelton,
  F., Miller, L., Simens, M., Askell, A., Welinder, P., Christiano, P., Leike,
  J., and Lowe, R. (2022).
\newblock Training language models to follow instructions with human feedback.
\newblock {\em arXiv preprint arXiv:2203.02155}.

\bibitem[Panch et~al. 2019]{panch2019artificial}
Panch, T., Pearson-Stuttard, J., Greaves, F., and Atun, R. (2019).
\newblock Artificial intelligence: opportunities and risks for public health.
\newblock {\em The Lancet Digital Health}, 1(1):e13--e14.

\bibitem[Paolanti et~al. 2018]{paolanti2018machine}
Paolanti, M., Romeo, L., Felicetti, A., Mancini, A., Frontoni, E., and
  Loncarski, J. (2018).
\newblock Machine learning approach for predictive maintenance in industry 4.0.
\newblock In {\em 2018 14th IEEE/ASME International Conference on Mechatronic
  and Embedded Systems and Applications (MESA)}, pages 1--6. IEEE.

\bibitem[Papakyriakopoulos et~al. 2020]{Papakyriakopoulos2020Bias}
Papakyriakopoulos, O., Hegelich, S., Serrano, J. C.~M., and Marco, F. (2020).
\newblock Bias in word embeddings.
\newblock In {\em Proceedings of the 2020 Conference on Fairness,
  Accountability, and Transparency}, FAT* '20, page 446–457. Association for
  Computing Machinery.

\bibitem[Phakhounthong et~al. 2018]{phakhounthong2018predicting}
Phakhounthong, K., Chaovalit, P., Jittamala, P., Blacksell, S.~D., Carter,
  M.~J., Turner, P., Chheng, K., Sona, S., Kumar, V., Day, N.~P., et~al.
  (2018).
\newblock Predicting the severity of dengue fever in children on admission
  based on clinical features and laboratory indicators: application of
  classification tree analysis.
\newblock {\em BMC Pediatrics}, 18:1--9.

\bibitem[Pivetta et~al. 2023]{pivetta2023}
Pivetta, M. V.~L., Simon, A.~H., Costa, M.~M., Abel, M., and Carbonera, J.~L.
  (2023).
\newblock A systematic evaluation of machine learning approaches for petroleum
  production forecasting.
\newblock In {\em IEEE 35th International Conference on Tools with Artificial
  Intelligence (ICTAI)}, pages 768--774. IEEE.

\bibitem[Prates et~al. 2020]{prates2020assessing}
Prates, M.~O., Avelar, P.~H., and Lamb, L.~C. (2020).
\newblock Assessing gender bias in machine translation: a case study with
  google translate.
\newblock {\em Neural Computing and Applications}, 32:6363--6381.

\bibitem[Radford et~al. 2021]{radford_learning_2021}
Radford, A., Kim, J.~W., Hallacy, C., Ramesh, A., Goh, G., Agarwal, S., Sastry,
  G., Askell, A., Mishkin, P., Clark, J., Krueger, G., and Sutskever, I.
  (2021).
\newblock Learning transferable visual models from natural language
  supervision.
\newblock In {\em Proceedings of the 38th International Conference on Machine
  Learning}, pages 8748--8763. {PMLR}.
\newblock {ISSN}: 2640-3498.

\bibitem[Rahmanifard and Plaksina 2019]{rahmanifard2019application}
Rahmanifard, H. and Plaksina, T. (2019).
\newblock Application of artificial intelligence techniques in the petroleum
  industry: a review.
\newblock {\em Artificial Intelligence Review}, 52(4):2295--2318.

\bibitem[Rajpurkar et~al. 2022]{rajpurkar2022ai}
Rajpurkar, P., Chen, E., Banerjee, O., and Topol, E.~J. (2022).
\newblock {AI} in health and medicine.
\newblock {\em Nature Medicine}, 28(1):31--38.

\bibitem[Ramos et~al. 2018]{Ramos+2018trc}
Ramos, G. {\relax de}.~O., Bazzan, A. L.~C., and {\relax da}~Silva, B.~C.
  (2018).
\newblock Analysing the impact of travel information for minimising the regret
  of route choice.
\newblock {\em Transportation Research Part C: Emerging Technologies},
  88:257--271.

\bibitem[Ribeiro et~al. 2021]{ribeiro2021robotic}
Ribeiro, J., Lima, R., Eckhardt, T., and Paiva, S. (2021).
\newblock Robotic process automation and artificial intelligence in industry
  4.0--a literature review.
\newblock {\em Procedia Computer Science}, 181:51--58.

\bibitem[Ribeiro et~al. 2016]{Ribeiro2016lime}
Ribeiro, M.~T., Singh, S., and Guestrin, C. (2016).
\newblock {"Why Should I Trust You?": Explaining the Predictions of Any
  Classifier}.
\newblock In {\em Proceedings of the 22nd {ACM} {SIGKDD} International
  Conference on Knowledge Discovery and Data Mining, San Francisco, CA, USA,
  August 13-17, 2016}, pages 1135--1144.

\bibitem[Roth et~al. 2022]{roth2022towards}
Roth, K., Pemula, L., Zepeda, J., Sch{\"o}lkopf, B., Brox, T., and Gehler, P.
  (2022).
\newblock Towards total recall in industrial anomaly detection.
\newblock In {\em Proceedings of the IEEE/CVF Conference on Computer Vision and
  Pattern Recognition}, pages 14318--14328.

\bibitem[Rumelhart et~al. 1986]{Rumelhart1986backprop}
Rumelhart, D.~E., Hinton, G.~E., and Williams, R.~J. (1986).
\newblock Learning representations by back-propagating errors.
\newblock {\em Nature}, 323(6088):533--536.

\bibitem[Russell et~al. 2015]{nature-ethics}
Russell, S., Hauert, S., Altman, R., and Veloso, M. (2015).
\newblock Ethics of artificial intelligence: Four leading researchers share
  their concerns and solutions for reducing societal risks from intelligent
  machines.
\newblock {\em Nature}, 521:415--418.

\bibitem[Russell and Norvig 2020]{Russell&Norvig2020}
Russell, S. and Norvig, P. (2020).
\newblock {\em Artificial Intelligence: {A} Modern Approach (4th Edition)}.
\newblock Pearson.

\bibitem[Santos and Bazzan 2021]{Santos&Bazzan2021}
Santos, G. D.~{\relax dos}. and Bazzan, A. L.~C. (2021).
\newblock Sharing diverse information gets driver agents to learn faster: an
  application in en route trip building.
\newblock {\em PeerJ Computer Science}, 7:e428.

\bibitem[Schuster and Paliwal 1997]{schuster1997bidirectional}
Schuster, M. and Paliwal, K.~K. (1997).
\newblock Bidirectional recurrent neural networks.
\newblock {\em IEEE Transactions on Signal Processing}, 45(11):2673--2681.

\bibitem[Schwalbe and Wahl 2020]{schwalbe2020artificial}
Schwalbe, N. and Wahl, B. (2020).
\newblock Artificial intelligence and the future of global health.
\newblock {\em The Lancet}, 395(10236):1579--1586.

\bibitem[Serradilla et~al. 2022]{serradilla2022deep}
Serradilla, O., Zugasti, E., Rodriguez, J., and Zurutuza, U. (2022).
\newblock Deep learning models for predictive maintenance: a survey,
  comparison, challenges and prospects.
\newblock {\em Applied Intelligence}, 52(10):10934--10964.

\bibitem[Sharir et~al. 2020]{sharir2020cost}
Sharir, O., Peleg, B., and Shoham, Y. (2020).
\newblock The cost of training nlp models: A concise overview.
\newblock {\em arXiv preprint arXiv:2004.08900}.

\bibitem[Silver et~al. 2017a]{Silver2017alphazero}
Silver, D., Hubert, T., Schrittwieser, J., Antonoglou, I., Lai, M., Guez, A.,
  Lanctot, M., Sifre, L., Kumaran, D., Graepel, T., et~al. (2017a).
\newblock Mastering chess and shogi by self-play with a general reinforcement
  learning algorithm.
\newblock {\em arXiv preprint arXiv:1712.01815}.

\bibitem[Silver et~al. 2017b]{Silver+2017}
Silver, D., Schrittwieser, J., Simonyan, K., Antonoglou, I., Huang, A., Guez,
  A., Hubert, T., Baker, L., Lai, M., Bolton, A., Chen, Y., Lillicrap, T., Hui,
  F., Sifre, L., van~den Driessche, G., Graepel, T., and Hassabis, D. (2017b).
\newblock Mastering the game of go without human knowledge.
\newblock {\em Nature}, 550(7676):354--359.

\bibitem[Stojanovic et~al. 2016]{stojanovic2016big}
Stojanovic, L., Dinic, M., Stojanovic, N., and Stojadinovic, A. (2016).
\newblock Big-data-driven anomaly detection in industry (4.0): An approach and
  a case study.
\newblock In {\em 2016 IEEE International Conference on Big Data}, pages
  1647--1652. IEEE.

\bibitem[Sutton and Barto 2018]{Sutton&Barto2018}
Sutton, R.~S. and Barto, A.~G. (2018).
\newblock {\em Reinforcement learning: An introduction}.
\newblock The MIT Press, second edition.

\bibitem[Thomas et~al. 2019]{Thomas2019seldonian}
Thomas, P.~S., Castro~da Silva, B., Barto, A.~G., Giguere, S., Brun, Y., and
  Brunskill, E. (2019).
\newblock Preventing undesirable behavior of intelligent machines.
\newblock {\em Science}, 366(6468):999--1004.

\bibitem[Toorajipour et~al. 2021]{toorajipour2021artificial}
Toorajipour, R., Sohrabpour, V., Nazarpour, A., Oghazi, P., and Fischl, M.
  (2021).
\newblock Artificial intelligence in supply chain management: A systematic
  literature review.
\newblock {\em Journal of Business Research}, 122:502--517.

\bibitem[Turing 1937]{Turing1937computability}
Turing, A.~M. (1937).
\newblock {On Computable Numbers, with an Application to the
  Entscheidungsproblem}.
\newblock {\em Proceedings of the London Mathematical Society},
  s2-42(1):230--265.

\bibitem[Turing 1950]{turing1950}
Turing, A.~M. (1950).
\newblock Computing machinery and intelligence.
\newblock {\em Mind}, 59(236):433--460.

\bibitem[Van~der Schaar et~al. 2021]{van2021artificial}
Van~der Schaar, M., Alaa, A.~M., Floto, A., Gimson, A., Scholtes, S., Wood, A.,
  McKinney, E., Jarrett, D., Lio, P., and Ercole, A. (2021).
\newblock How artificial intelligence and machine learning can help healthcare
  systems respond to {COVID-19}.
\newblock {\em Machine Learning}, 110:1--14.

\bibitem[Vaswani et~al. 2017]{vaswani2017attention}
Vaswani, A., Shazeer, N., Parmar, N., Uszkoreit, J., Jones, L., Gomez, A.~N.,
  Kaiser, {\L}., and Polosukhin, I. (2017).
\newblock Attention is all you need.
\newblock {\em Advances in Neural Information Processing Systems}, 30.

\bibitem[Vinyals et~al. 2019]{Vinyals+2019}
Vinyals, O., Babuschkin, I., Czarnecki, W., Mathieu, M., Dudzik, A., Chung, J.,
  Choi, D., Powell, R., Ewalds, T., Georgiev, P., Oh, J., Horgan, D., Kroiss,
  M., Danihelka, I., Huang, A., Sifre, L., Cai, T., Agapiou, J., Jaderberg, M.,
  and Silver, D. (2019).
\newblock Grandmaster level in {StarCraft II} using multi-agent reinforcement
  learning.
\newblock {\em Nature}, 575:350–--354.

\bibitem[von Neumann 1956]{von1956probabilistic}
von Neumann, J. (1956).
\newblock Probabilistic logics and the synthesis of reliable organisms from
  unreliable components.
\newblock {\em Automata Studies}, 34:43--98.

\bibitem[Wang and Luo 2019]{wang2019machine}
Wang, L. and Luo, M. (2019).
\newblock Machine learning applications and opportunities in ic design flow.
\newblock In {\em 2019 International Symposium on VLSI Design, Automation and
  Test (VLSI-DAT)}, pages 1--3. IEEE.

\bibitem[Warren et~al. 2023]{Warren2023}
Warren, D.~S., Dahl, V., Eiter, T., Hermenegildo, M.~V., Kowalski, R.~A., and
  Rossi, F., editors (2023).
\newblock {\em Prolog: The Next 50 Years}, volume 13900 of {\em Lecture Notes
  in Computer Science}.
\newblock Springer.

\bibitem[Watkins 1989]{Watkins1989}
Watkins, C. (1989).
\newblock {\em Learning from Delayed Rewards}.
\newblock PhD thesis, University of Cambridge.

\bibitem[Watkins and Dayan 1992]{Watkins&Dayan1992}
Watkins, C. J. C.~H. and Dayan, P. (1992).
\newblock {Q}-learning.
\newblock {\em Machine Learning}, 8(3):279--292.

\bibitem[Wei et~al. 2019]{Wei+2019colight}
Wei, H., Li, Z., Xu, N., Zhang, H., Zheng, G., Zang, X., Chen, C., Zhang, W.,
  Zhu, Y., and Xu, K. (2019).
\newblock Colight: Learning network-level cooperation for traffic signal
  control.
\newblock In {\em Proceedings of the 28th ACM International Conference on
  Information and Knowledge Management}, pages 1913--1922. Association for
  Computing Machinery.

\bibitem[Wiering 2000]{Wiering2000}
Wiering, M. (2000).
\newblock Multi-agent reinforcement learning for traffic light control.
\newblock In {\em Proceedings of the Seventeenth International Conference on
  Machine Learning ({ICML} 2000)}, pages 1151--1158.

\bibitem[Wong et~al. 2021]{wong2021external}
Wong, A., Otles, E., Donnelly, J.~P., Krumm, A., McCullough, J.,
  DeTroyer-Cooley, O., Pestrue, J., Phillips, M., Konye, J., Penoza, C., et~al.
  (2021).
\newblock External validation of a widely implemented proprietary sepsis
  prediction model in hospitalized patients.
\newblock {\em JAMA Internal Medicine}, 181(8):1065--1070.

\bibitem[{World Health Organization} 2021]{world2021ethics}
{World Health Organization} (2021).
\newblock Ethics and governance of artificial intelligence for health: {WHO}
  guidance.

\bibitem[Xu et~al. 2019]{xu2019translating}
Xu, J., Yang, P., Xue, S., Sharma, B., Sanchez-Martin, M., Wang, F., Beaty,
  K.~A., Dehan, E., and Parikh, B. (2019).
\newblock Translating cancer genomics into precision medicine with artificial
  intelligence: applications, challenges and future perspectives.
\newblock {\em Human Genetics}, 138(2):109--124.

\bibitem[Yau et~al. 2017]{Yau+2017}
Yau, K.-L.~A., Qadir, J., Khoo, H.~L., Ling, M.~H., and Komisarczuk, P. (2017).
\newblock A survey on reinforcement learning models and algorithms for traffic
  signal control.
\newblock {\em ACM Comput. Surv.}, 50(3).

\bibitem[Yu et~al. 2018]{yu2018artificial}
Yu, K.-H., Beam, A.~L., and Kohane, I.~S. (2018).
\newblock Artificial intelligence in healthcare.
\newblock {\em Nature Biomedical Engineering}, 2(10):719--731.

\bibitem[Zhang et~al. 2019]{zhang2019food}
Zhang, X., Zhou, T., Zhang, L., Fung, K.~Y., and Ng, K.~M. (2019).
\newblock Food product design: a hybrid machine learning and mechanistic
  modeling approach.
\newblock {\em Industrial \& Engineering Chemistry Research},
  58(36):16743--16752.

\bibitem[Zhu et~al. 2021]{zhu2021demand}
Zhu, X., Ninh, A., Zhao, H., and Liu, Z. (2021).
\newblock Demand forecasting with supply-chain information and machine
  learning: Evidence in the pharmaceutical industry.
\newblock {\em Production and Operations Management}, 30(9):3231--3252.

\bibitem[Zipfel et~al. 2023]{zipfel2023anomaly}
Zipfel, J., Verworner, F., Fischer, M., Wieland, U., Kraus, M., and Zschech, P.
  (2023).
\newblock Anomaly detection for industrial quality assurance: A comparative
  evaluation of unsupervised deep learning models.
\newblock {\em Computers \& Industrial Engineering}, 177:109045.

\end{thebibliography}
\end{small}

\end{document}